\documentclass[letterpaper, 10 pt, conference]{ieeeconf}

\IEEEoverridecommandlockouts

\usepackage[T1]{fontenc}

\usepackage[noadjust]{cite}
\usepackage{graphicx}
\usepackage{subfig}
\usepackage{bm}
\usepackage{amsmath, xparse}
\usepackage{amssymb}
\usepackage{algorithm}
\usepackage[noend]{algpseudocode}
\usepackage{lipsum}
\usepackage{footnote}
\usepackage{etoolbox}
\usepackage{array}
\usepackage{hyperref}

\newtoggle{extended_version}
\toggletrue{extended_version}

\renewcommand{\baselinestretch}{0.945}

\graphicspath{{figures/}}

\AddToHook{cmd/@maketitle/after}{\ADDINITIALFIGURE}

\newcommand{\ADDINITIALFIGURE}{
\begin{minipage}{2.0\columnwidth}
    \expandafter\def\csname @captype\endcsname{figure}
    \subfloat[]{\includegraphics[width=0.24\columnwidth]{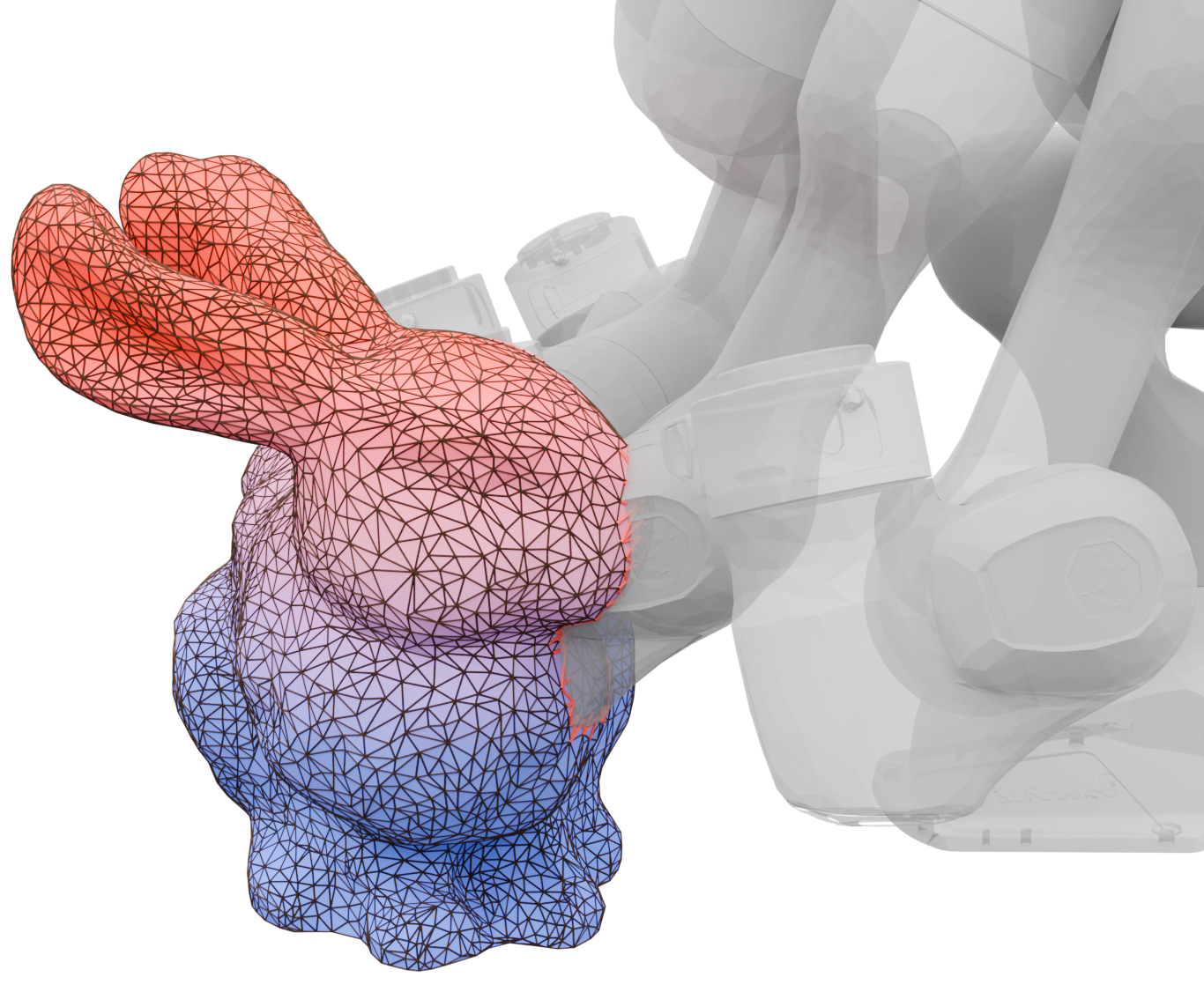}\label{fig: headline_d}}\ 
    \subfloat[]{\includegraphics[width=0.24\columnwidth]{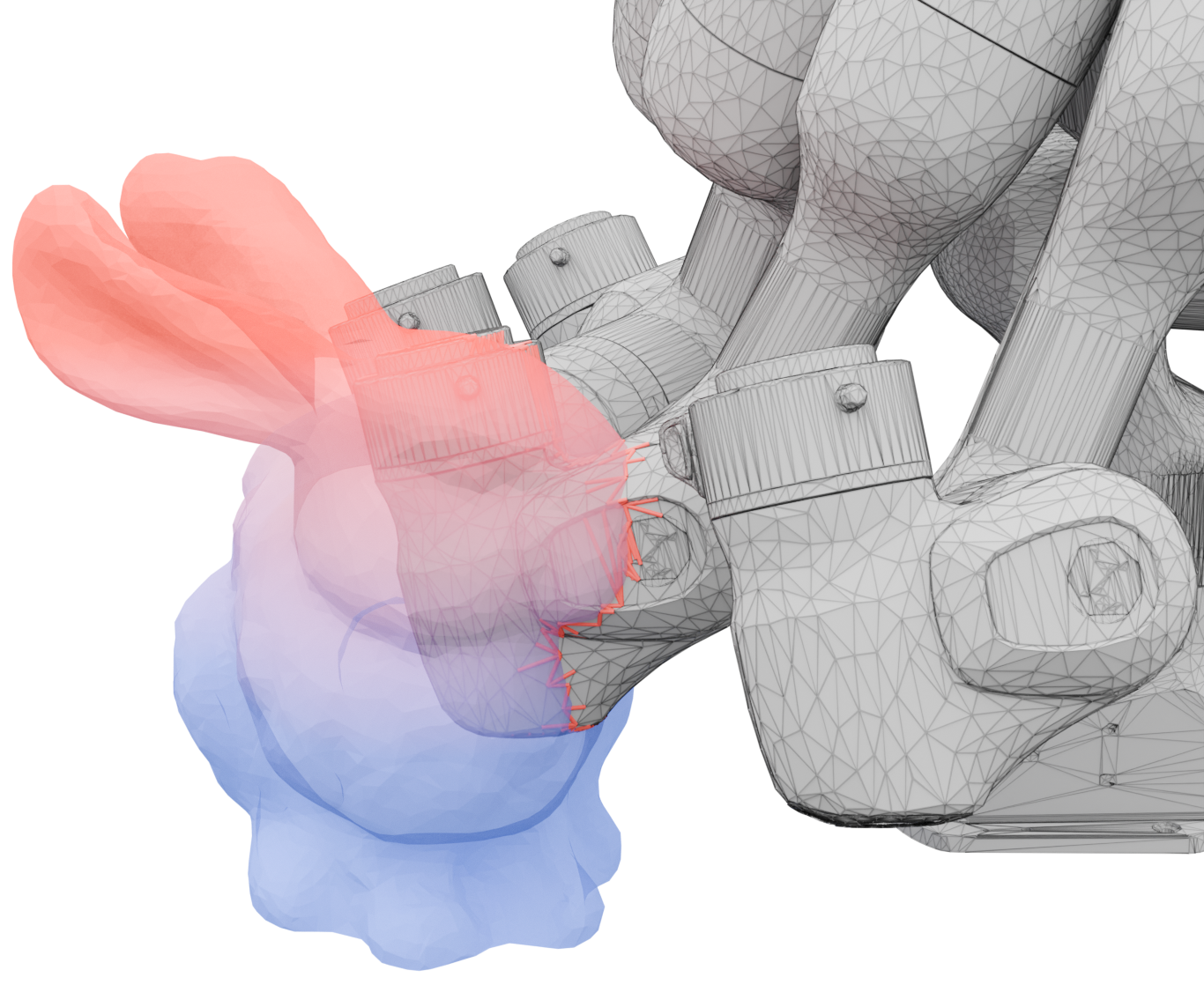}\label{fig: headline_r}}\
    \subfloat[]{\includegraphics[width=0.24\columnwidth]{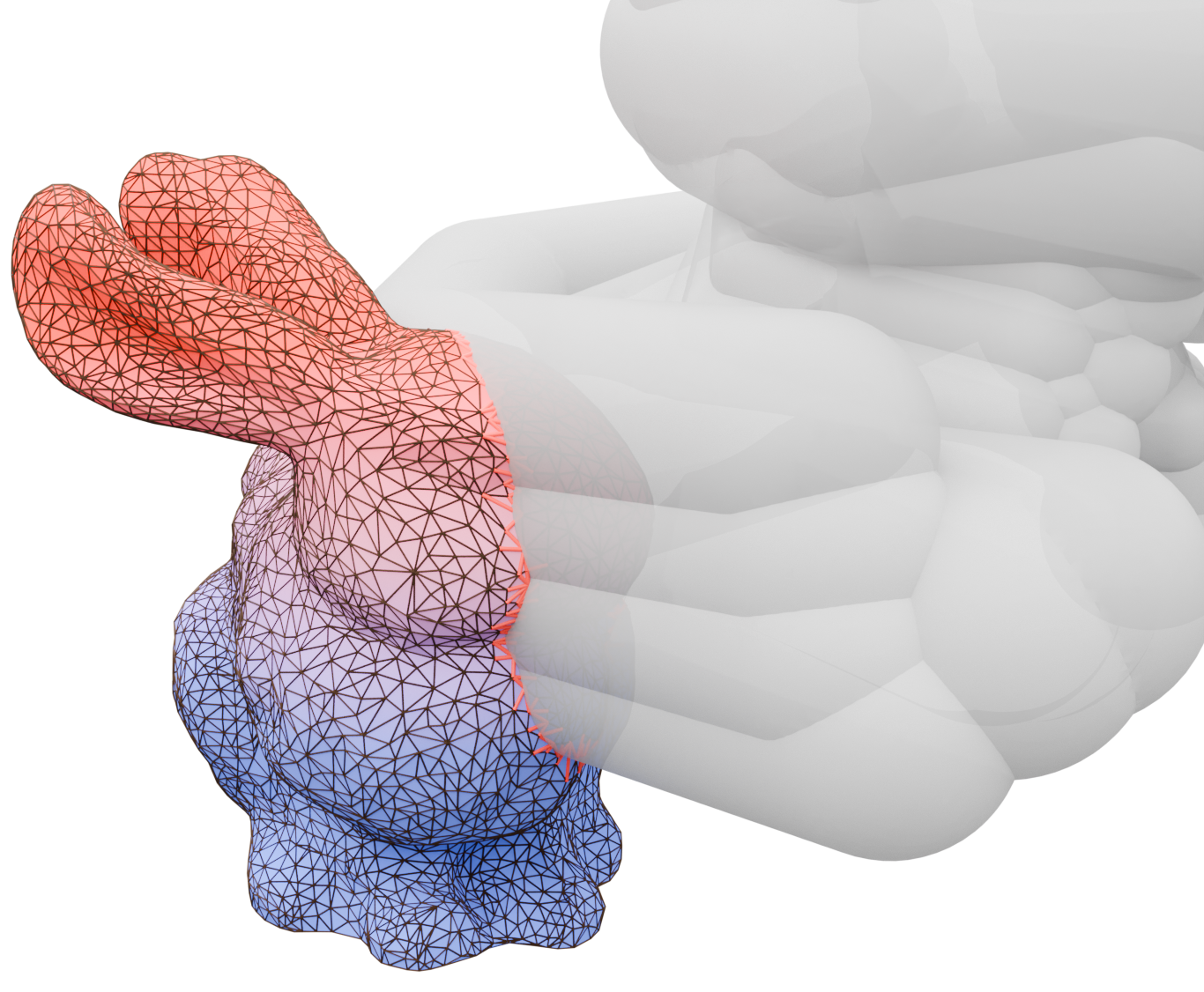}\label{fig: headline_l}}\ 
    \subfloat[]{\includegraphics[width=0.24\columnwidth]{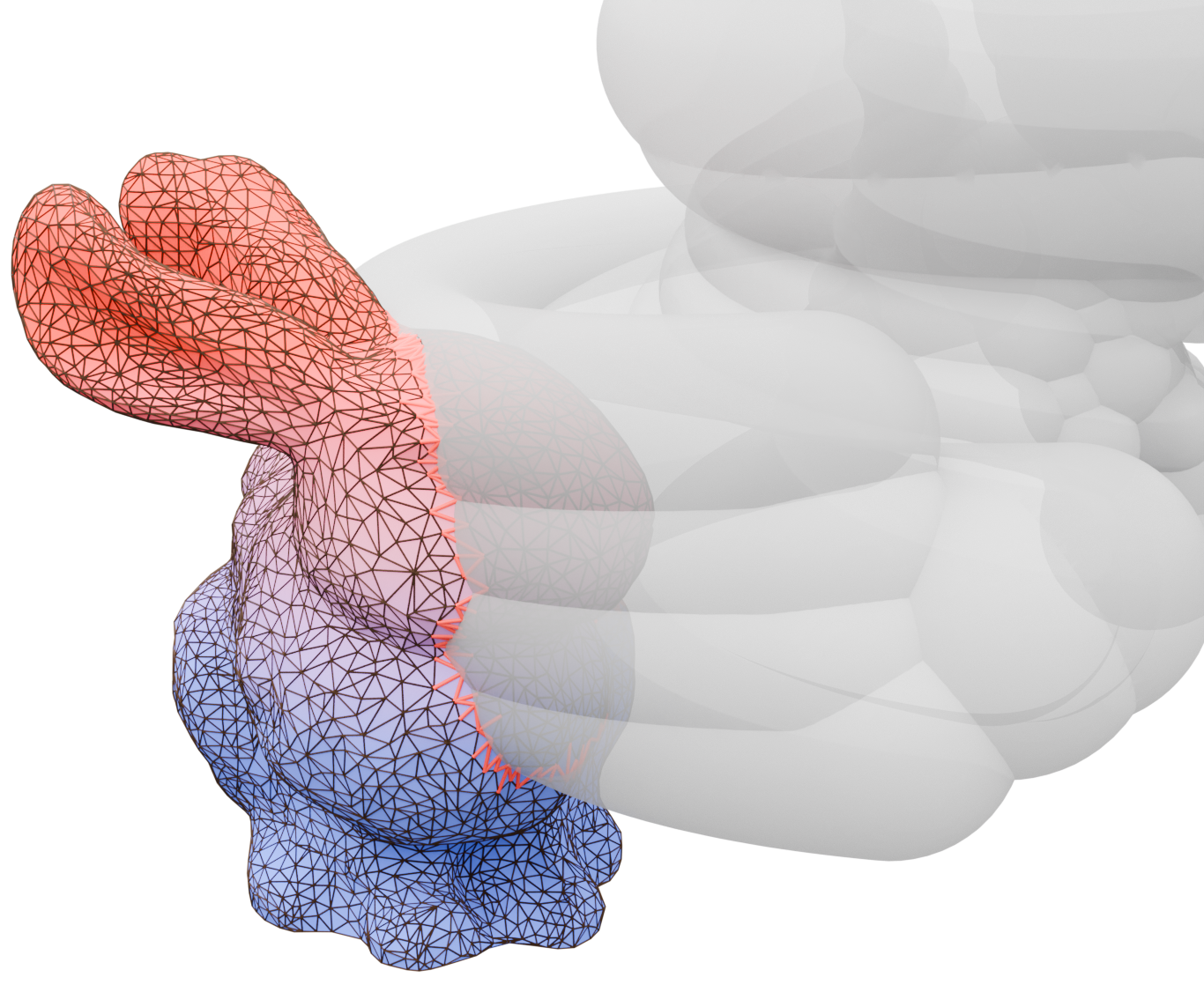}\label{fig: headline_q}}\
    \vspace{-2pt}
    \caption{Demonstration of our collision detection methods: discrete-pose collision detection by ray-tracing (a) along obstacle meshes  and (b) along robot meshes, and continuous collision detection by ray-tracing against swept sphere-approximated robot volumes (c) along piecewise-linear paths or (d) along quadratic B-spline paths.}
    \setcounter{figure}{1}
    \label{fig: headline}
    \vspace{-10pt}
\end{minipage}}

\begin{document}

\iftoggle{extended_version}{

\begin{minipage}{2.0\columnwidth}
\begin{center}
\textbf{Note Regarding ICRA Publication}
\end{center}

This article is an extended version of a paper accepted at the IEEE International Conference on Robotics and Automation (ICRA 2025).

The final published version is available in the ICRA 2025 Proceedings via \href{https://ieeexplore.ieee.org/abstract/document/11128528}{IEEE Xplore}.

\textcopyright 2025 IEEE. Personal use of this material is permitted. Permission from IEEE must be obtained for all other uses, in any current or future media, including reprinting/republishing this material for advertising or promotional purposes, creating new collective works, for resale or redistribution to servers or lists, or reuse of any copyrighted component of this work in other works.

The new and extended portions of this arXiv preprint are \textcopyright 2025 by the authors.
\end{minipage}
}{}

\setlength{\textfloatsep}{0pt}

\title{\LARGE \bf Hardware-Accelerated Ray Tracing for Discrete and Continuous Collision Detection on GPUs\vspace{-7pt}}

\author{Sizhe Sui$^{1}$, Luis Sentis$^{1}$, and Andrew Bylard$^{2}$\vspace{5pt}
\thanks{$^{1}$Sizhe Sui is a Dexterity fellow and he and Luis Sentis are with the Department of Aerospace Engineering and Engineering Mechanics, The University of Texas at Austin, USA, {\tt\small \{sizhe.sui, lsentis\}@utexas.edu}}
\thanks{$^{2}$Andrew Bylard is with Dexterity, Inc., Redwood City, CA, USA, {\tt\small andrew.bylard@dexterity.ai}}
\iftoggle{extended_version}{\thanks{Source code at https://github.com/Ssz990220/RTCollisionDetection}}{\thanks{Extended version at https://arxiv.org/abs/2409.09918, source code at https://github.com/Ssz990220/RTCollisionDetection}}
}

\maketitle

\begin{abstract}

This paper presents a set of simple and intuitive robot collision detection algorithms that show substantial scaling improvements for high geometric complexity and large numbers of collision queries by leveraging hardware-accelerated ray tracing on GPUs. It is the first leveraging hardware-accelerated ray-tracing for direct volume mesh-to-mesh discrete collision detection and applying it to continuous collision detection. We introduce two methods: Ray-Traced Discrete-Pose Collision Detection for exact robot mesh to obstacle mesh collision detection, and Ray-Traced Continuous Collision Detection for robot sphere representation to obstacle mesh swept collision detection, using piecewise-linear or quadratic B-splines. For robot link meshes totaling 24k triangles and obstacle meshes of over 190k triangles, our methods were up to 2.8 times faster in batched discrete-pose queries than a state-of-the-art GPU-based method using a sphere robot representation. For the same obstacle mesh scene, our sphere-robot continuous collision detection was up to 7 times faster depending on trajectory batch size. We also performed detailed measurements of the volume coverage accuracy of various sphere/mesh pose/path representations to provide insight into the tradeoffs between speed and accuracy of different robot collision detection methods.

\end{abstract}

\section{Introduction}
Collision detection is a fundamental building block of robot motion planning. However, it is a complex operation with two major scaling issues. The first is the high geometric complexity of robots and obstacles. Real-world objects often have complex geometry, and accurate modeling of geometric details is often useful to reduce conservatism. Reducing conservatism is itself key to improving solution optimality or in many cases finding solutions at all. For $N$ robot and $M$ obstacle geometry primitives, naive collision detection, not including self-collisions, has time complexity $O(MN)$. This can be improved to $O(\log M \log N)$ via two-way bounding-volume hierarchy (BVH) traversal ~\cite{bergen1997efficient,gottschalk1996obbtree,larsen2000fast,klosowski1998efficient}, but this still may be too slow if the constant factor is large.

The second scaling issue is the large number of collision detection queries needed for planning algorithms, often on the order of thousands of pose checks in addition to sub-path/edge checks. Additionally, a robot performing complex tasks may perform hundreds of motion planning calls when evaluating candidate tasks and ways to perform each task. Parallelism from CPU multithreading or SIMD instructions (as in~\cite{thomason2023MotionsMicrosecondsVectorized, Ramsey2024SIMD}) saturates very quickly. GPUs scale much better, able to execute simultaneous instructions from thousands of times more threads. However, GPU-based collision detection~\cite{pan2012gpu} is difficult to implement---high-performance BVH construction and traversal on the GPU is very complex---leading to few well-maintained software implementations.

Another factor is that CPU cores and previously-used GPU cores are general-purpose, not specialized for collision detection related tasks. However, with the introduction of hardware-accelerated ray tracing (Sec.~\ref{sec: HART}), this may no longer be the case. Hardware-accelerated ray-tracing accelerates ray-triangle intersection and ray-to-BVH traversal, and ray-tracing engines implement accelerated BVH construction, accelerating and abstracting away operations very relevant to robot collision detection, resulting both in faster and simpler GPU algorithm implementations. However, ray-tracing cores have just begun to be explored for collision-related queries~\cite{bylard2021, mandarapu2024mochi}.

This paper leverages hardware-accelerated ray tracing for novel collision detection algorithms, achieving precise and fast collision detection and demonstrating the potential of hardware-accelerated ray tracing for applications like robot motion planning. Our main contributions are as follows:
\begin{enumerate}
\item \textbf{Ray-Traced Discrete-Pose Collision Detection (RT-DCD)}: Performs simple and intuitive mesh-to-mesh pose collision detection up to 2.8x faster than GPU-based mesh-to-sphere counterpart cuRobo~\cite{sundaralingam2023CuRoboParallelizedCollisionFree} and up to 27x faster than CPU-based FCL~\cite{pan2012fcl}, e.g., checking 4,096 articulated-arm poses against a $>$190k triangle mesh collision scene in 4 ms.
\item \textbf{Ray-Traced Continuous Collision Detection (RT-CCD)}: Constructs swept volumes of sphere-approximated robots along trajectories using OptiX curves and spheres and detects collisions with mesh obstacles. Performs up to 7x faster than cuRobo, depending on trajectory batch size and curve type (piecewise linear vs. quadratic B-spline).
\item \textbf{Volume Coverage Accuracy Pipeline}: Uses high-resolution GPU-based voxelization to measure the volume coverage accuracy of different robot representations (sphere vs. mesh) for discrete poses and for trajectory approximations (e.g., discretized, piecewise linear, quadratic B-spline). We applied this in addition to standard collision detection accuracy results for obstacle-scene-independent insight into the tradeoffs between speed and accuracy of different approaches.
\end{enumerate}

\section{Background and Related Works}

Robot boolean collision detection can be split into two problems: discrete-pose collision detection (DCD), which checks for collisions on single robot poses, and continuous collision detection (CCD), which checks for collisions on full continuous robot paths. These problems can be further broken down into direct collision detection in 3D Euclidean space versus constructing and checking for collision constraints in the robot configuration space (C-space). The former method uses forward kinematics to compute robot link poses during a collision query and checks for overlap between links and obstacles in 3D space. The latter method may compute forward kinematics during collision constraint construction, but not during a query as it can then check C-space directly to find collision-free space.

Collision checking becomes more time-consuming as the geometric complexity of the robot and obstacles increases. It is quite common to simplify geometric representations of the robot and/or obstacles (e.g., using primitive shapes such as spheres or capsules), but this may be too conservative in cluttered scenes or tight spaces. Broad-phase/narrow-phase methods can also improve computational scaling through quick elimination of groups of collision geometries via different types of bounding volume hierarchies (BVH), such as AABB~\cite{bergen1997efficient}, OBB~\cite{gottschalk1996obbtree}, SSV~\cite{larsen2000fast}, and k-DOP~\cite{klosowski1998efficient}.

\subsection{CPU-Based DCD Methods}
The Flexible Collision Library (FCL)~\cite{pan2012fcl} collision checks in 3D space with BVH acceleration and has seen perhaps the most widespread adoption in robotics, valued for its support for many geometry, collision pair, and query types. VAMP~\cite{thomason2023MotionsMicrosecondsVectorized} and CAPT~\cite{Ramsey2024SIMD} further accelerated collision detection of sphere robots through CPU SIMD instructions. 
However, these libraries are CPU-bound, and CPU multithreading, SIMD, and BVH scaling are often not enough to handle both the geometric complexity and quantity of collision queries needed for low-conservatism real-time planning.

Methods which construct and query collision constraints directly in C-space face challenges such as the curse of dimensionality and the often highly nonlinear mapping between robot configuration and the robot's occupancy of 3D space, often leading to complex collision-free/in-collision C-space sets which are difficult to fit explicit representations to.
Methods like Fastron~\cite{das2020LearningBasedProxyCollision} and CollisionGP~\cite{munoz2023CollisionGPGaussianProcessBased} use data-driven approaches to approximate collision-free C-space, and Motion Planning Networks~\cite{qureshi2019MotionPlanningNetworks} embed collision detection in the joint space in a neural network. However, the statistical nature of these methods complicates accuracy guarantees, compromising safety in some applications.

\subsection{GPU-Based DCD Methods without Ray Tracing}

Pushing past the limits of CPU-bound methods, some works have turned to parallel computation on GPUs. Pan, et al.~\cite{pan2012gpu} designed a GPU-based parallel BVH traversal algorithm that maximizes GPU execution efficiency through load balancing.
Our work provides both simplification and speedup using BVH operations abstracted away and hardware-accelerated in ray-tracing engines.
CuRobo~\cite{sundaralingam2023CuRoboParallelizedCollisionFree} uses parallel point-signed distance queries on the GPU to detect collisions with complex obstacle scenes, but this requires a conservative sphere-based approximation of the robot, whereas we perform exact detection on robot meshes.

Some recent methods like PairwiseNet~\cite{kim2023PairwiseNetPairwiseCollision} and ClearanceNet~\cite{chasekew2021NeuralCollisionClearance} leverage neural networks and GPUs for higher throughput, but ClearanceNet only achieved 91-96\% detection accuracy and PairwiseNet requires training on each mesh pair for collision detection, whereas our DCD method detects all penetrating collisions between arbitrary meshes with minimal precomputation time.

\subsection{CCD Methods}

In addition to checking individual poses, robot motion planning algorithms must check for collision-free paths through continuous collision detection (CCD), also known as swept-volume collision detection. Paths can be discretized into individual poses and checked using DCD algorithms, but this can result in missed collisions or even robots tunneling through obstacles~\cite{ERICSON20057}. However, the explicit construction of swept-volume geometries is very complex. Sellán, et al.~\cite{sellan2021SweptVolumesSpacetime} overcame this by extending the zero signed distance surface in 4D space-time, while Zhang, et al.~\cite{zhang2023ContinuousImplicitSDF} and Wang, et al.~\cite{wang2024ImplicitSweptVolume} described swept volumes implicitly using generalized semi-infinite programming (GSIP). However, these methods require seconds to minutes to build a single swept volume, making them unsuitable for real-time collision detection.

Conservative advancement (CA) methods~\cite{lin1993efficient, tang2014HierarchicalControlledAdvancement, tang2011CCQEfficientLocal} offer an alternative by computing a motion bound to determine the maximum forward sweep time step on a trajectory. The cuRobo library~\cite{sundaralingam2023CuRoboParallelizedCollisionFree} uses a modified bisection and CA method on robots approximated by spheres, achieving CCD with excellent batch performance through parallel execution on GPUs. Our research proposes a new CCD method for sphere-approximated robots, using a GPU-based ray-tracing method to check against explicitly constructed swept volumes and achieving faster speeds depending on trajectory batch size and higher accuracy through swept curves which are higher order than cuRobo's method allows.

\subsection{Hardware-Accelerated Ray Tracing} \label{sec: HART}
Ray tracing is a technique primarily used in computer graphics to render images by simulating light as it propagates, reflects, scatters, and refracts in a scene. This requires finding accurate intersections between rays and often detailed, high-dimensional objects to produce realistic visual effects. Due to the large number of required rays and complex ray computations (traversing BVH trees, narrow-phase intersection tests, secondary ray scattering, etc.), ray tracing was for decades not feasible for real-time applications. However, recent GPU architectures include cores dedicated to ray tracing, making real-time ray tracing a reality. 

We have previously explored hardware-accelerated ray tracing for computing signed distances (which can serve as collision detection), finding contact points, and formulating trajectory optimization constraints between robot and obstacle meshes~\cite{bylard2021}.
The Mochi engine~\cite{mandarapu2024mochi} explored applications to DCD between spheres, triangle meshes, and generic geometries which can be AABB-bounded and have a point-containment function. However, its mesh DCD does not treat meshes as volumes (it cannot detect an obstacle fully contained in a robot mesh or vice versa), and it has suboptimalities for common robot DCD applications (for example, not distinguishing desired robot/obstacle collision pairs and not exploring the best batching strategies for multi-query multi-link robot applications).

Our work is the first leveraging hardware-accelerated ray tracing for direct volume mesh-to-mesh discrete collision detection (rather than indirect through signed-distance measurement) and the first applying it to continuous collision detection. To showcase the potential of adapting hardware-accelerated ray tracing for collision detection, we use NVIDIA ray-tracing (RT) cores, which include hardware acceleration for ray-traced BVH traversal and ray-triangle intersections, and NVIDIA's OptiX~\cite{Parker2010optix} ray-tracing engine as the platform for implementing our algorithms. These algorithms can also transfer to other ray-tracing frameworks and hardware platforms which provide similar functionality.

\begin{figure}[t!]
	\centering\includegraphics[width=\columnwidth]{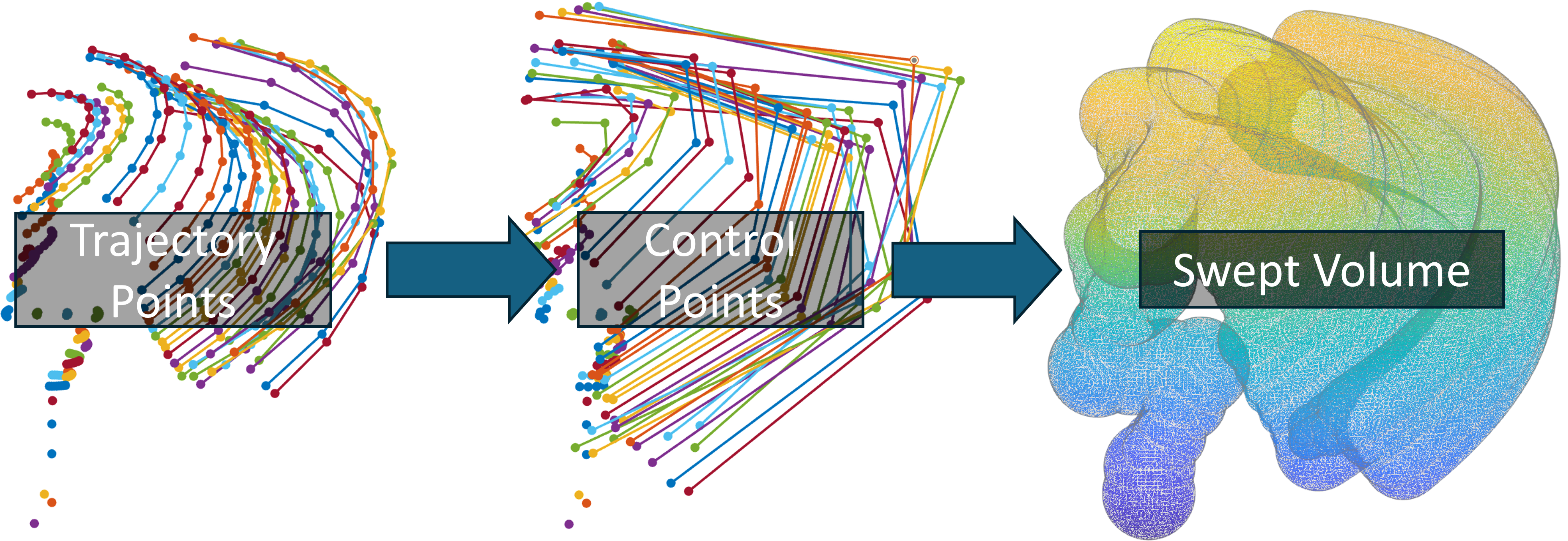}
        \caption{Process of generating swept robot volumes for CCD from piecewise-linear or quadratic/cubic B-spline curves.}\label{fig: curve_gen}
\end{figure}
\section{Methodology}

\subsection{Ray Tracing Volume Mesh-to-Mesh Collision Detection}\label{sec: RTCD}

First, we introduce a method to detect collisions between a pair of mesh volumes using ray tracing. Our proposed method designates one mesh as the source of the rays and the other as the target. We trace rays along the triangle edges of the source triangle mesh towards the target mesh. These rays are along the surface of the source mesh, so any ray intersection indicates a collision between the two meshes.

However, tracing rays along the surface of only one of the meshes is insufficient to capture all potential collisions. For example, if the target mesh is needle-thin, it may just penetrate the center of a source mesh triangle, resulting in a missed detection. Most real-world objects do not have thin protruding features, but this is also a risk when the target mesh has large triangles. For example, a rectangular surface ideally consists of two large triangles instead of many smaller ones. To address this issue, we also trace rays along the triangle edges of the other mesh and aggregate the results.

A further corner case is when the robot is fully contained inside an obstacle or vice versa, which is common in aerial, underwater, or space motion planning scenes with unanchored robots / obstacles. For these cases, it is sufficient to have watertight meshes with correctly oriented triangles (normals facing out) and to trace one maximum-length ray from the interior of each mesh toward all target meshes. If the ray hits more triangle back faces than front (which can be detected through OptiX), the ray origin is in collision. Note that this also handles target meshes that overlap. 

These tracing strategies produce exact volume mesh-to-mesh detection of penetrating collisions, which is sufficient for most robot motion planning applications.

\subsection{Discrete-Pose Collision Detection}\label{sec: DCD Alg}

We developed three approaches to detect collisions between robot and obstacle volume meshes: \mbox{ObsToRobot} (Fig.\ref{fig: headline_d}) traces rays along obstacles toward robot links, \mbox{RobotToObs} (Fig.\ref{fig: headline_r}) traces rays along robot links posed by forward kinematics toward obstacles, and ``two-way'' simply combines the results of \mbox{ObsToRobot} and \mbox{RobotToObs} for exact detection results. Alg.~\ref{alg: discrete CD} shows the inputs of each, including a robot kinematics / link mesh model and a mesh obstacle scene. The output is a boolean array indicating whether each robot configuration is in collision with an obstacle. Each algorithm is GPU-parallelized to batch operations such as forward kinematics across robot configurations and aggregating results across rays, and each uses the method in Sec. \ref{sec: RTCD} to identify in-collision mesh pairs.

\begin{algorithm}[t!]
	\caption{Discrete pose collision detection (RT-DCD)}
	\label{alg: discrete CD}
	\textbf{Data:} \text{Robot configurations} $\{\theta_i\}$, \text{robot model}, \text{scene}, \text{obb$_\text{o}$}
 \begin{algorithmic}[1]
        \State $\bm{T_f}\gets$ \text{BatchForwardKinematics}($\{\theta_i\}$,\text{ robot})
        \State \text{obb$_\text{r}$}$\gets$\text{BatchTransformRobotOBBs}($\bm{T_f}$,\text{ robot})
        \State \text{mask}$\gets$\text{BatchOBBDetect}(\text{obb$_\text{o}$},\text{ obb$_\text{r}$})
        \State \text{links$_\text{m}$}$\gets$\text{BatchCompactLinks}(\text{mask},\text{ robot})
        \If{\textnormal{ObsToRobot}}
            \State \Return \text{BatchLaunchObstacleRays}(\text{scene},\text{ links$_\text{m}$})
        \ElsIf{\textnormal{RobotToObs}}
            \State $\bm{T}_{\bm{f}\text{m}}\gets$\text{BatchCompactLinkPoses}(\text{mask},\text{ robot},$\;\bm{T_f}$)
            \State \Return \text{BatchLaunchLinkRays}(\text{scene},\text{ links$_\text{m}$},$\;\bm{T}_{\bm{f}\text{m}}$)
        \EndIf
\end{algorithmic}
\end{algorithm}

For RobotToObs, we trace different rays for each robot configuration. However, for ObsToRobot, we trace against batches of robot configurations at the same time by placing posed pre-built link mesh BVHs in an instance acceleration structure (IAS), an OptiX BVH which can reuse other BVHs for scenes with geometries duplicated in different poses.

\begin{figure}[t!]
    \centering\includegraphics[width=1.0\columnwidth]{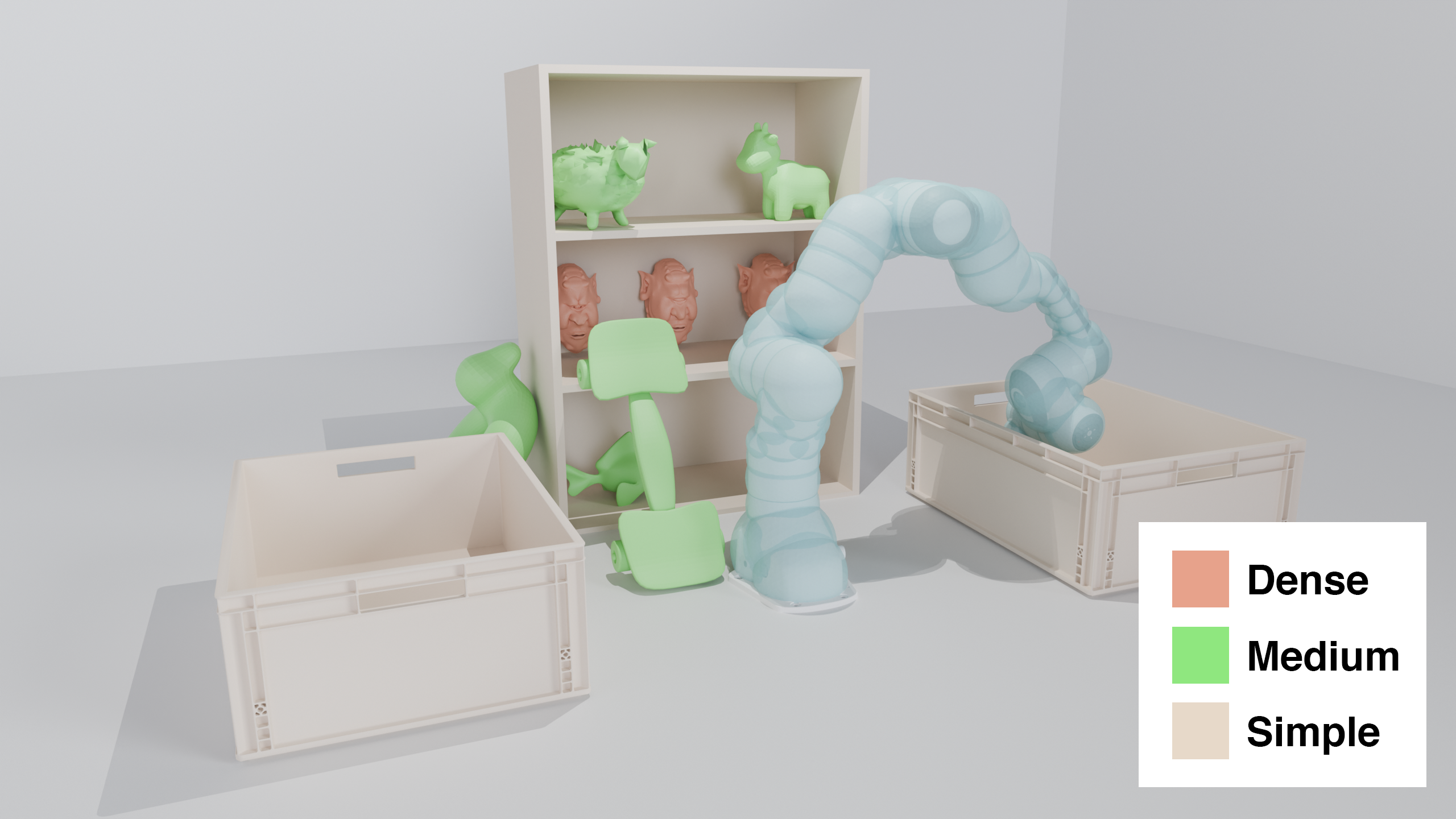}
    \caption{Collision scene for detection benchmarks. Each level of complexity also contains the meshes of previous levels.}\label{fig: collision scene}
\end{figure}

To increase speedup, we precompute an oriented bounding box (OBB) for each obstacle and robot mesh via a simple PCA method taking $\sim$300us for a mesh with 20,000 vertices~\cite{dimitrov2009bounds}. We then use these OBBs for broad-phase collision detection (i.e. single-level BVHs for simplicity), eliminating a large number of unnecessary posed links from IASs and rays from tracing steps. For example, in the scene shown in Fig~\ref{fig: collision scene}, only 1/4th of the associated OBBs are in collision. To aggregate useful posed links and rays from broadphase result masks, we use stream compaction~\cite{Pharr2005GPU} a parallel algorithm that gathers subsets of data of interest. Although stream compaction has poor scaling compared other common classes of parallel algorithms on the GPU, we found it was more than worth it to reduce inputs to ray-tracing steps, which tend to take by far the majority of compute time.

\subsection{Continuous Collision Detection} \label{subsec: CCD}

We also developed a method for continuous collision detection for robots approximated by spheres. In addition to triangles, OptiX allows representing geometries by spheres and curves. OptiX curves, usually used to render hair, grass, etc., are 3D paths with a flat or round (circular) cross-section. These paths can be defined in a variety of ways: piecewise linear, uniform quadratic / cubic B-splines, Catmull-Rom splines, or B\'ezier curves. Curves may also be given a constant- or varying-radius cross-section along its path. In this work, we leverage the fact that OptiX curves with a constant radius and a sphere at each endpoint are equivalent to the swept volume of a sphere of the same radius. 

For piecewise-linear swept volumes, we use batch forward kinematics to compute robot link sphere centers on poses along a path, linearly interpolating sphere centers between subsequent poses. For more accurate path approximations, we also implemented quadratic and cubic B-splines \cite{farin2014curves}, for which we compute control points from path waypoints via least squares as in \cite{dong2017efficient}.

Since the number of control points determines the least-squares pseudo-inverse matrix, each control point becomes a simple constant-weighted sum of all trajectory points. Thus we store this pseudo-inverse matrix in GPU constant memory for efficient broadcasting across instances of a parallel reduction kernel used to compute the curve control points across all links and pose paths.

One challenge is that OptiX meshes, spheres, and curves are all considered hollow, but for curves (and spheres starting in OptiX 9.0), ray traces against them cannot detect back-face hits (intersections starting from a curve's interior). To address this, we choose ray directions along the edges of each obstacle mesh such that they form a strongly-connected directed graph. This guarantees catching vertex collisions as any out-of-collision vertex must have a ray-directed path to any in-collision vertex. We precompute these graphs using a simple, greedy approach: ensure each triangle has edge rays forming a consistent orientation (e.g., clockwise), first by growing regions of alternating triangle orientations where possible. For remaining triangles, if there are already three edge rays, we choose the orientation matching two of the rays. Otherwise we set a random orientation.  This takes 30 ms to compute for an example 40k triangle mesh, resulting in only $\sim$10\% of edges being traced in both directions.

\section{Experiments and Results} \label{sec: Exp&Results}
In this section, we evaluate the volume coverage accuracy of various collision body representations as well as the computational performance and collision-detection accuracy of the proposed algorithms compared to other approaches, with the Flexible Collision Library (FCL)~\cite{pan2012fcl} as a CPU baseline and the cuRobo library (with CUDA graphs enabled and self-collision disabled)~\cite{sundaralingam2023CuRoboParallelizedCollisionFree} providing comparisons to state-of-the-art GPU-based methods. We did not compare to Mochi~\cite{mandarapu2024mochi} as the library is not open-sourced and adds unnecessary complexity for our application (see Sec.~\ref{subsec:dcd_accuracy}).

\begin{figure}[t!]
    \centering\includegraphics[width=1.0\columnwidth]{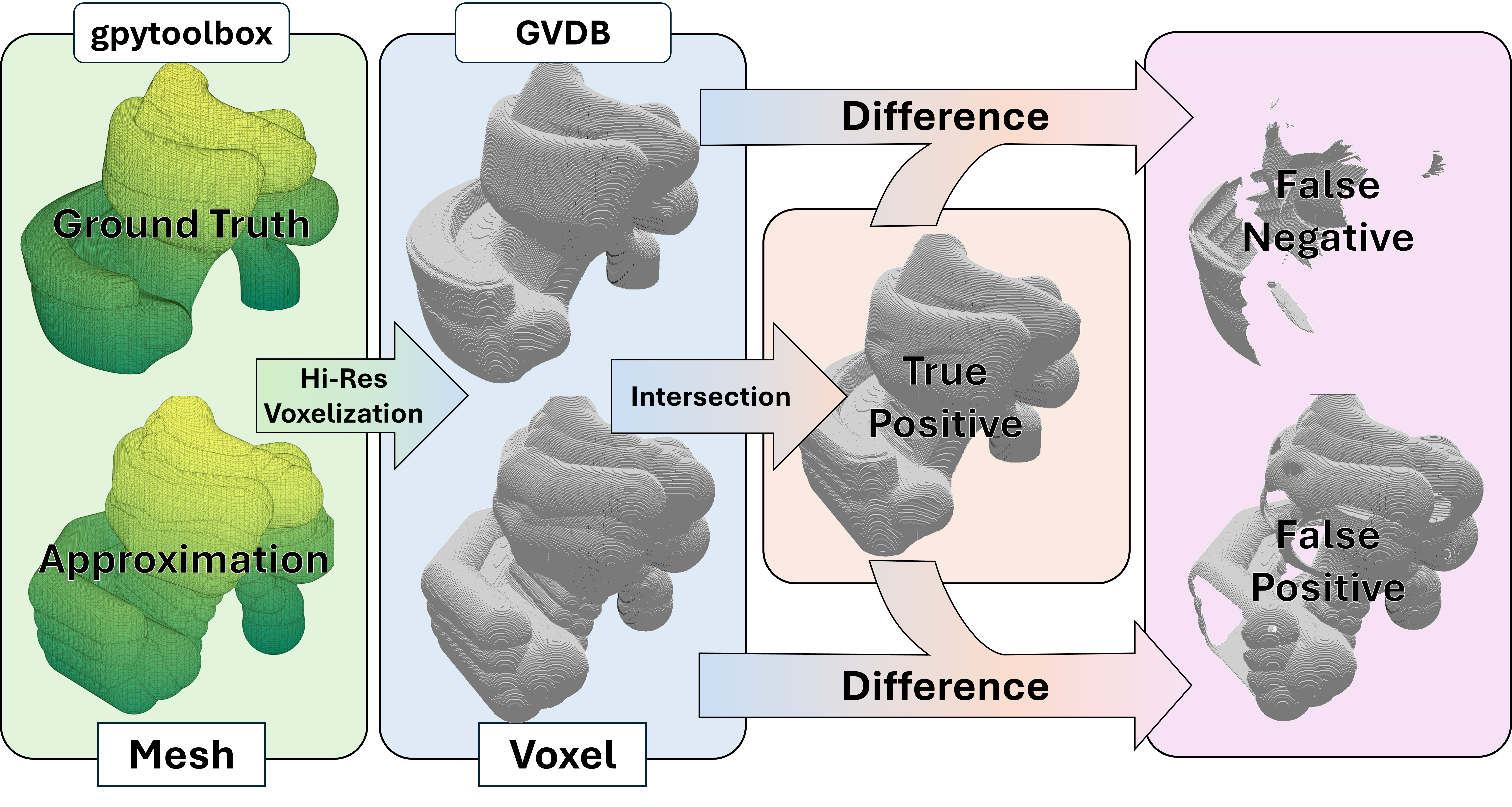}
    \caption{Workflow for measuring the accuracy of a representation format, depicting an exaggerated case. The false positive/negative volumes are not used in our accuracy metrics but are shown here for illustration.}\label{fig: acc_measure}
    \vspace{5pt}
\end{figure}

\subsection{Robot and Collision Environments} \label{sec: environment}

We used a Franka Panda robot for all experiments, adapting its mesh from the official visual mesh model, adjusted to be watertight and simplified to around 24k~/~11k triangles~/~edges (mostly replacing very small triangles with little effect on accuracy). For a sphere representation, we used 62 spheres to achieve the volume coverage accuracy (defined in Sec.~\ref{sec: AccMea}) reported in the first column of Table~\ref{table: acc}. We used the same mesh model for comparisons with FCL and the same sphere model for comparisons with cuRobo.

\newcolumntype{C}{>{\centering\arraybackslash}p{0.8cm}}
{\renewcommand{\arraystretch}{1.2}
\setlength{\tabcolsep}{4pt}
\begin{table*}[t!]
\centering
\begin{tabular}{|c|c|c||c|c|c|c|c|c|c|c|c|C|C|}
\hline
\multicolumn{1}{|l|}{} & \multicolumn{2}{c||}{Discrete Pose} & \multicolumn{5}{c|}{Discretized Trajectory} & \multicolumn{4}{c|}{Piecewise Linear} & \multicolumn{2}{c|}{Quadratic Spline} \\
\hline
\multicolumn{1}{|c|}{Robot link representation} & \multicolumn{1}{c|}{Sphere} & \multicolumn{1}{c||}{Mesh} & \multicolumn{5}{c|}{Mesh} & \multicolumn{4}{c|}{Sphere} & \multicolumn{2}{c|}{Sphere} \\
\hline
Number of control points & 1 & 1 & 2 & 4 & 8 & \textbf{16} & \textbf{32} & 2 & 4 & \textbf{8} & \textbf{16} & 4 & \textbf{8} \\
\hline
Precision (\%) & 77.40 & 100.00 & 100.00 & 100.00 & 100.00 & \textbf{100.00} & \textbf{100.00} & 72.22 & 84.70 & \textbf{85.61} & \textbf{85.39} & 85.08 & \textbf{85.30} \\
\hline
Recall rate (\%) & 99.53 & 100.00 & 47.74 & 69.09  & 85.63  & \textbf{93.65} & \textbf{96.25}  & 73.07 & 97.15 & \textbf{99.90} & \textbf{99.99} & 99.90 & \textbf{99.99} \\

\hline
Detection false positives (\%) & 2.79 & 0.00 & 0.00 & 0.00 & 0.00 & \textbf{0.00} & \textbf{0.00} & 1.66 & 0.90 & \textbf{1.14} & \textbf{1.39} & 1.35 & \textbf{0.93} \\
\hline
Detection false negatives (\%) & 0.00 & 0.07 & 19.59 & 3.74 & 1.07 & \textbf{0.27} & \textbf{0.13} & 15.84 & 1.21 & \textbf{0.10} & \textbf{0.00} & 0.42 & \textbf{0.00} \\
\hline

\end{tabular}
\caption{Accuracy results for different discrete pose and continuous trajectory collision approximation strategies.}
\label{table: acc}
\vspace{-11pt}
\end{table*}
}

\begin{figure*}[t!]
    \centering
    \subfloat[Dense Scene]{\includegraphics[width=0.32\textwidth]{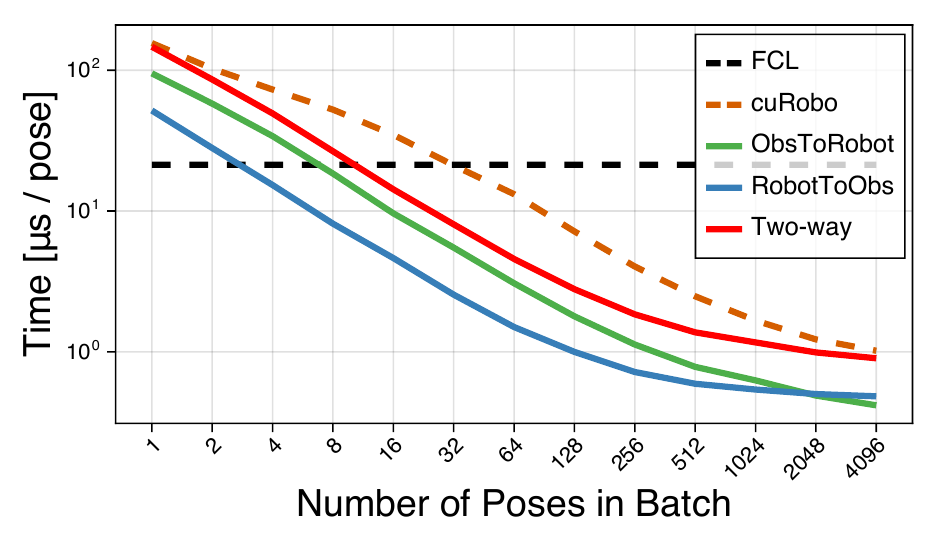}}\
    \subfloat[Medium Scene]{\includegraphics[width=0.32\textwidth]{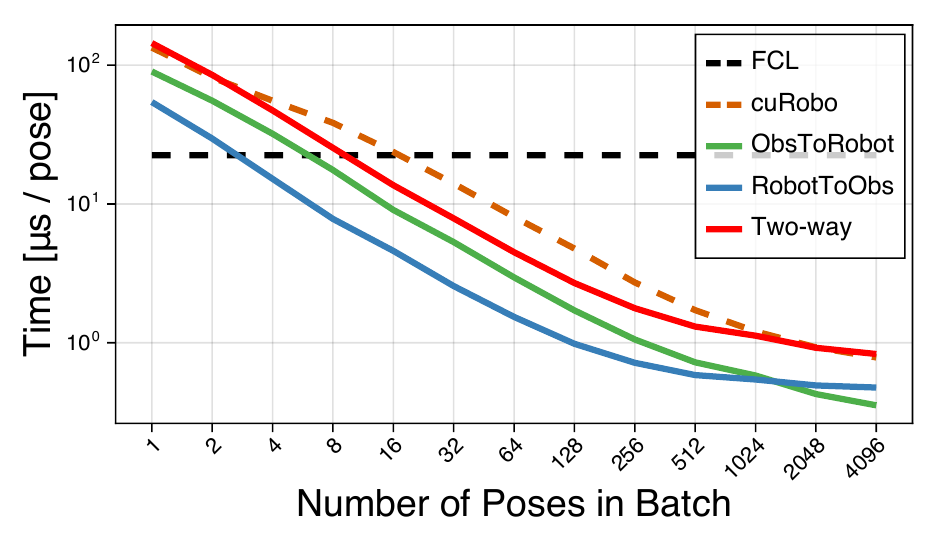}\label{fig: discrete shelf}}\
    \subfloat[Simple Scene]{\includegraphics[width=0.32\textwidth]{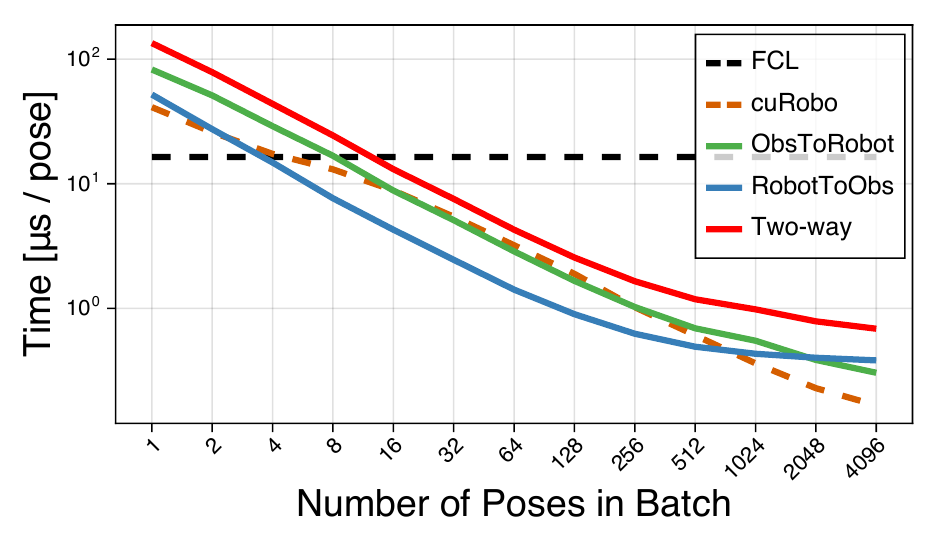}\label{fig: discrete simple shelf}}\
        \caption{Runtime for all tested discrete collision detection (DCD) algorithms in different scenes.}\label{fig: discrete result}
        \vspace{-11pt}
\end{figure*}

\begin{figure*}[t!]
    \centering
    \subfloat[Dense Scene]{\includegraphics[width=0.32\textwidth]{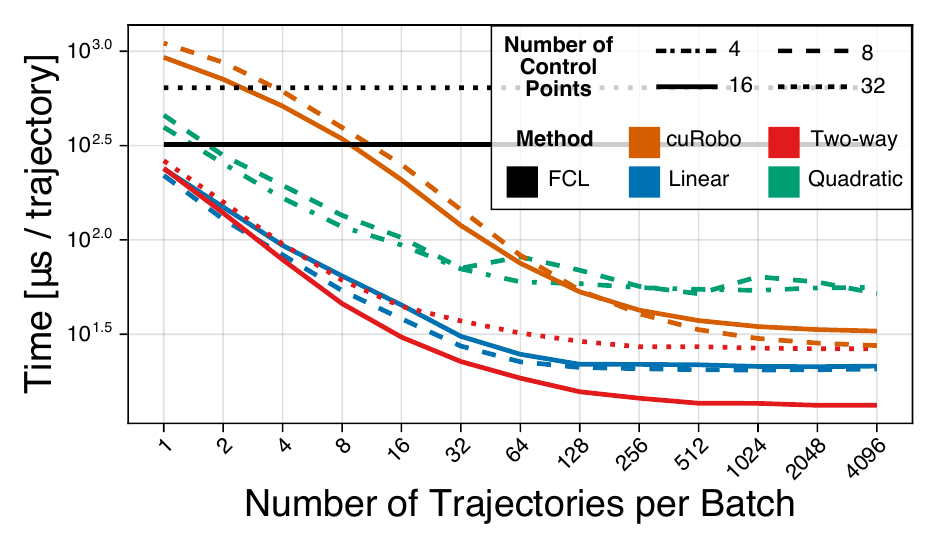}\label{fig: curve dense shelf}}\
    \subfloat[Medium Scene]{\includegraphics[width=0.32\textwidth]{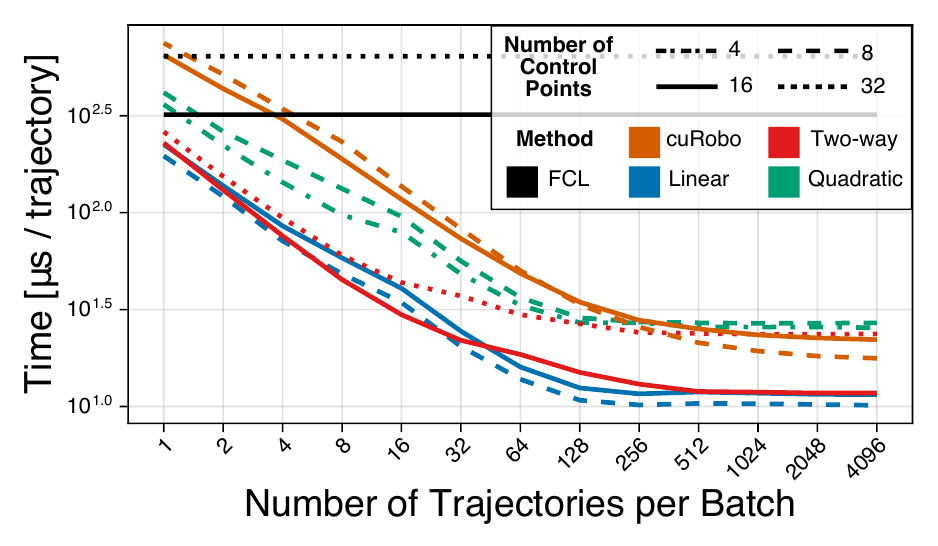}\label{fig: curve shelf}}\
    \subfloat[Simple Scene]{\includegraphics[width=0.32\textwidth]{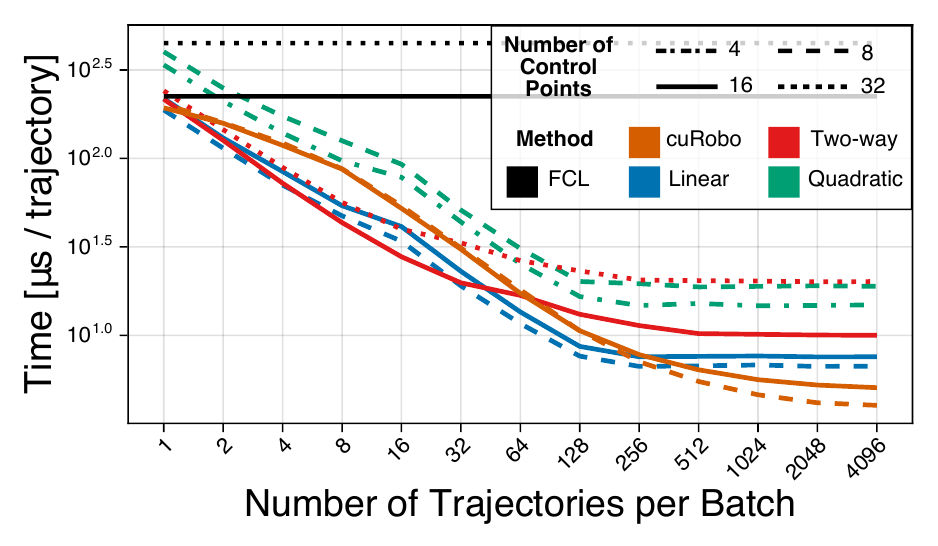}\label{fig: curve simple shelf}}\
        \caption{Runtime for all tested continuous collision detection (CCD) algorithms in different scenes.}\label{fig: curve result}
        \vspace{-20pt}
\end{figure*}

To benchmark computational speed and collision-checking accuracy of our method, we constructed several versions of a collision scene, adding obstacle meshes~\cite{crane2013robust} to increase complexity as shown in Fig.~\ref{fig: collision scene}. The simple, medium, and dense versions of the scene consisted of over 15k~/~22k, 71k~/~107k, and 191k~/~321k triangles~/~edges, respectively. We performed all experiments on an AMD Ryzen 7 7700x CPU and an RTX 4070 Super GPU. We wrote the library in C++/CUDA with OptiX 8.0, compiling with g++13 and CUDA 12.6, and we pre-allocated all GPU memory buffers to avoid dynamic allocations on critical compute paths.

\subsection{Volume-Coverage Accuracy Measurement}\label{sec: AccMea}
Various works use different volume representations (implicit or explicit) of discrete robot poses and continuous (swept) robot pose paths to perform boolean collision detection. Though collision detection is the end objective, it is difficult to get a holistic measurement of the accuracy of these volume representations through just collision detection experiments as the results depend heavily on the selected poses and set of obstacles being checked against.

Thus, we aim to supplement collision-detection accuracy tests with detailed volume-coverage accuracy measurements. To do this, we implemented a custom pipeline (illustrated in Figure~\ref{fig: acc_measure}) using high-resolution voxels to measure the overlap and discrepancies between the representations and the ground truth, defined by watertight volume meshes of the robot links. For discrete poses, we used Trimesh~\cite{trimesh} to compute the union of all link meshes, posed by forward kinematics. For swept volumes, we used gpytoolbox~\cite{gpytoolbox} to perform the method in \cite{sellan2021SweptVolumesSpacetime} for constructing a watertight swept mesh of each link mesh, merging these into a watertight robot swept mesh. We converted non-mesh robot volume representations such as spheres into meshes to use the pipeline.

We measured volume coverage accuracy through two metrics: precision and recall. \textbf{Precision} is the true positive volume (overlap between the approximated and true volumes) over the total approximated volume. Higher is better, meaning that the approximation is less conservative. \textbf{Recall} is the true positive volume over the total true volume. Again, higher is better, meaning that the approximation is safer, not missing volume that could be in collision.

Measuring the true positive volume requires a suitable representation of the volume intersection between the true and approximated robot meshes. Unfortunately, although each mesh is watertight, constructing corresponding watertight mesh intersections is challenging. Instead, we approximated volume intersection through high-resolution dense voxelization of our volumes using GVDB \cite{hoetzlein2016GVDBRaytracingSparse}, a GPU-based hierarchical voxel processing library which allowed scaling our experiments to millions of voxels with a minimum volume of 1 mm$^3$ for accurate volume statistics. 

To generate discrete test robot poses, we used quasi-random Halton sampling~\cite{halton1964Algo} in C-space. To generate test pose trajectories, we then sampled 30 trajectories of varying lengths connecting Halton-sampled joint configuration pairs via straight lines in C-space. The resulting volume accuracy results are detailed in Table~\ref{table: acc}. Cubic spline results are omitted as they showed minimal accuracy improvement over the quadratic spline while significantly increasing runtime.

The discretized mesh approximation of swept volumes achieved perfect precision but low recall rate until high numbers of discrete poses per trajectory, as may be expected. The sphere-based swept volumes achieved high recall rate at 8 control points for piecewise linear approximations and just 4 for quadratic B-splines. However, the conservative sphere approximation resulted in a max precision of around 85\%.

\subsection{Discrete Pose Collision Detection Evaluation} \label{sec: discrete result}
We tested our discrete-pose collision detection methods on a set of 4096 Halton-sampled robot poses, checking against the three versions of the scene as described in Sec.~\ref{sec: environment}. To measure the effect of batch sizes on speed, we tested our methods and cuRobo's on power-of-2 batch sizes up to 4096 (we measured the FCL CPU baseline at a batch size of 1). We included all steps of Alg. \ref{alg: discrete CD} in benchmarking, including forward kinematics on the GPU.

Fig.~\ref{fig: discrete result} shows speed results, including separate results for the ObsToRobot/RobotToObs RT-DCD methods and both combined for the highest accuracy (two-way). Despite performing a much more complex collision check than cuRobo (mesh-to-mesh vs. mesh-to-sphere), our methods outpaced cuRobo by up to 2.8x on medium to dense scenes, scaling past CPU performance at smaller batch sizes. The cuRobo method catches up to our two-way method at thousands of poses per batch, but such large batch sizes are often not needed in motion planning (e.g., PRM can use one large batch, but most planners use waves of smaller batches).

The cuRobo method does perform faster in simple scenes. This is because (1) our mesh-based methods check many more primitives (up to $>$35k robot edge rays even on the simple scene vs. 62 spheres), and (2) the hardware-accelerated BVH traversal we use scales better with denser meshes (compared to the warp library underlying cuRobo, which using a single normal CUDA thread for each point-to-mesh BVH traversal). The cost of (1) is not sufficiently diminished by our simple broadphase method or offset by (2) on ``simple'' scenes. This can be addressed through improved broadphase or conservative simplification of the meshes rays are traced along. However, as the meshes traced against get more detailed, the scaling of hardware-accelerated BVH traversal do lead to higher speeds overall.

\subsection{Continuous Collision Detection Evaluation}\label{sec: continuous result}
For CCD, we benchmarked our piecewise-linear and quadratic B-spline sphere swept volume RT-CCD against discretized mesh pose trajectories checked with our two-way RT-DCD and cuRobo, the only other GPU-based library implementing CCD. Due to cuRobo's sphere-based CA-style algorithm, its swept volumes are identical to our piecewise linear curves. Table~\ref{table: acc} shows detection accuracy results, and Fig.~\ref{fig: curve result} shows speed for different trajectory batch sizes and trajectory control points. We again omit our cubic B-spline results (see Sec.~\ref{sec: AccMea}). Our piecewise-linear and discretized methods outpaced cuRobo on all except large batch sizes on simple scenes, with discretization being the fastest on dense scenes at the cost of recall rate. Quadratic B-splines had the best accuracy at slower speeds, but were still faster than cuRobo at most batch sizes for medium/dense scenes.

\section{Discussion}\label{sec: discussion}
\subsection{Accuracy Analysis}

\subsubsection{RT-DCD Accuracy}\label{subsec:dcd_accuracy}
As mentioned, our two-way volume mesh-to-mesh RT-DCD catches all penetrating collisions. However, it cannot detect mere surface contact between meshes as NVIDIA's RT cores do not guarantee catching ray-triangle intersections where the ray is coplanar with the triangle or starts/ends at the triangle surface (such intersections have little use in graphics rendering). Mochi \cite{mandarapu2024mochi} quadruples the number of triangles to catch more of such collisions, but it still cannot catch all of them (such as triangles or edges fully contained in a coplanar triangle) without tracing longer rays with more complex post-processing logic. Since surface contact with zero penetration can often be omitted in collision-avoidance (often subject in any case to numerical precision issues that one can avoid with slight robot or obstacle inflation), we are content instead to detect only penetrating collisions with a much simpler approach.

\subsubsection{RT-CCD Accuracy}
Unlike two-way RT-DCD, our RT-CCD methods are not exact as they only detect swept sphere collisions with mesh edges, not faces. For a swept sphere with radius $R$ and a triangle face whose maximum inscribed-circle radius is $r$, the maximum penetration depth is $R - \sqrt{R^2-r^2}$ if $R > r$ and infinite otherwise when not accounting for neighboring swept spheres. A solution is to split large triangles into smaller ones until the maximum inscribed-circle radius across the whole mesh drops the maximum penetration depth below some acceptable limit.

\subsubsection{Numerical Error}
NVIDIA RT cores only allow up to single-precision floating point geometry and ray definitions, and only the highest-end NVIDIA GPUs include double-precision CUDA arithmetic units (double- to single-precision theoretical performance ratio is 1:2 on the H100 but is 1:64 on the RTX 4090). Numerical error can thus add up through geometry definition, forward kinematics, and ray tracing, in our case leading to the 0.07\% RT-DCD false negative rate in Table~\ref{table: acc}. If missed penetrations on the order of single-precision numerical error are unacceptable for an application, users can add some slight inflation to the robot mesh.

\subsection{Performance of RT-CCD}\label{sec: ccd analysis}
It is notable that 16-pose discretized trajectories checked with our mesh-to-mesh RT-DCD is faster than even our piecewise-linear mesh-to-swept-sphere RT-CCD on dense scenes (though the discretized trajectory suffers expected recall rate deficiencies). The slowness of our RT-CCD compared to RT-DCD has a number of contributing factors.

First, ray tracing against curves, particularly high-order curves, is much more expensive than against triangles as up through the Ada generation GPUs used in this work, the narrow-phase intersection is not hardware-accelerated, running instead on normal CUDA cores. Thus it only benefits from hardware-acceleration during BVH traversal. This is expected to improve in Blackwell generation GPUs, which hardware-accelerate narrow-phase ray-traces against piecewise-linear curves and spheres.

Second, our RT-CCD requires expensive curve BVHs builds from scratch on the critical path, unlike RT-DCD which can reuse precomputed mesh BVHs in an IAS. Finally, curve BVHs are less efficient than for triangle meshes as the sphere extrusions lead to large AABBs, the bounding volume used in current NVIDIA RT cores. This is often addressed by splitting long curves into smaller segments or across multiple smaller AABBs, but this leads to longer BVH build times on our critical path which we found overall gives no speedup.
\section{Conclusion}
In this paper, we presented novel algorithms for discrete and continuous robot collision detection using hardware-accelerated ray tracing, showing impressive scaling for high geometric complexity and number of parallel queries. This is a clear demonstration of the value of hardware-accelerated ray-tracing as a new tool for fast, high-complexity, and high-throughput collision/geometric queries in robotics. In the future, we will explore more ray-traced collision queries such as self-collision and point-cloud collision detection.

\bibliographystyle{IEEEtran}
\bibliography{reference}

\begin{thebibliography}{10}
\providecommand{\url}[1]{#1}
\csname url@samestyle\endcsname
\providecommand{\newblock}{\relax}
\providecommand{\bibinfo}[2]{#2}
\providecommand{\BIBentrySTDinterwordspacing}{\spaceskip=0pt\relax}
\providecommand{\BIBentryALTinterwordstretchfactor}{4}
\providecommand{\BIBentryALTinterwordspacing}{\spaceskip=\fontdimen2\font plus
\BIBentryALTinterwordstretchfactor\fontdimen3\font minus \fontdimen4\font\relax}
\providecommand{\BIBforeignlanguage}[2]{{%
\expandafter\ifx\csname l@#1\endcsname\relax
\typeout{** WARNING: IEEEtran.bst: No hyphenation pattern has been}%
\typeout{** loaded for the language `#1'. Using the pattern for}%
\typeout{** the default language instead.}%
\else
\language=\csname l@#1\endcsname
\fi
#2}}
\providecommand{\BIBdecl}{\relax}
\BIBdecl

\bibitem{bergen1997efficient}
G.~van~den Bergen, ``Efficient collision detection of complex deformable models using {AABB} trees,'' \emph{Journal of Graphics Tools}, vol.~2, no.~4, pp. 1--13, 1997.

\bibitem{gottschalk1996obbtree}
S.~Gottschalk, M.~C. Lin, and D.~Manocha, ``{OBBTree}: A hierarchical structure for rapid interference detection,'' in \emph{SIGGRAPH: International Conference on Computer Graphics and Interactive Techniques}, 1996.

\bibitem{larsen2000fast}
E.~Larsen, S.~Gottschalk, M.~C. Lin, and D.~Manocha, ``Fast distance queries with rectangular swept sphere volumes,'' in \emph{IEEE {{International Conference}} on {{Robotics}} and {{Automation}} ({{ICRA}})}, 2000.

\bibitem{klosowski1998efficient}
J.~T. Klosowski, M.~Held, J.~S. Mitchell, H.~Sowizral, and K.~Zikan, ``Efficient collision detection using bounding volume hierarchies of {k-DOPs},'' \emph{IEEE Transactions on Visualization and Computer Graphics}, vol.~4, no.~1, pp. 21--36, 1998.

\bibitem{thomason2023MotionsMicrosecondsVectorized}
W.~Thomason, Z.~Kingston, and L.~E. Kavraki, ``Motions in microseconds via vectorized sampling-based planning,'' in \emph{{{IEEE International Conference}} on {{Robotics}} and {{Automation}} ({{ICRA}})}, 2024.

\bibitem{Ramsey2024SIMD}
C.~W. Ramsey, Z.~Kingston, W.~Thomason, and L.~E. Kavraki, ``{CAPT}: Collision-affording point trees: {SIMD}-amenable nearest neighbors for fast collision checking,'' in \emph{Robotics: Science and Systems (RSS)}, 2024.

\bibitem{pan2012gpu}
J.~Pan and D.~Manocha, ``{GPU-based} parallel collision detection for fast motion planning,'' \emph{The International Journal of Robotics Research (IJRR)}, vol.~31, no.~2, pp. 187--200, 2012.

\bibitem{bylard2021}
A.~Bylard, ``Leveraging the geometric structure of robotic tasks for motion design,'' Ph.D. dissertation, Stanford University, 2021.

\bibitem{mandarapu2024mochi}
D.~K. Mandarapu, N.~James, and M.~Kulkarni, ``Mochi: Fast \& exact collision detection,'' \emph{arXiv preprint arXiv:2402.14801}, 2024.

\bibitem{sundaralingam2023CuRoboParallelizedCollisionFree}
B.~Sundaralingam, S.~K.~S. Hari, A.~Fishman, C.~Garrett, K.~Van~Wyk, V.~Blukis, A.~Millane, H.~Oleynikova, A.~Handa, F.~Ramos, N.~Ratliff, and D.~Fox, ``{{CuRobo}}: Parallelized collision-free robot motion generation,'' in \emph{{{IEEE International Conference}} on {{Robotics}} and {{Automation}} ({{ICRA}})}, 2023.

\bibitem{pan2012fcl}
J.~Pan, S.~Chitta, and D.~Manocha, ``$\text{FCL}$: A general purpose library for collision and proximity queries,'' in \emph{IEEE International Conference on Robotics and Automation (ICRA)}, 2012.

\bibitem{das2020LearningBasedProxyCollision}
N.~Das and M.~Yip, ``Learning-based proxy collision detection for robot motion planning applications,'' \emph{IEEE Transactions on Robotics (T-RO)}, vol.~36, no.~4, pp. 1096--1114, Aug. 2020.

\bibitem{munoz2023CollisionGPGaussianProcessBased}
J.~Mu{\~n}oz, P.~Lehner, L.~E. Moreno, A.~{Albu-Sch{\"a}ffer}, and M.~A. Roa, ``{{CollisionGP}}: Gaussian process-based collision checking for robot motion planning,'' \emph{IEEE Robotics and Automation Letters (RA-L)}, vol.~8, no.~7, pp. 4036--4043, Jul. 2023.

\bibitem{qureshi2019MotionPlanningNetworks}
A.~H. Qureshi, A.~Simeonov, M.~J. Bency, and M.~C. Yip, ``Motion {{Planning Networks}},'' in \emph{IEEE {{International Conference}} on {{Robotics}} and {{Automation}} ({{ICRA}})}, 2019.

\bibitem{kim2023PairwiseNetPairwiseCollision}
J.~Kim and F.~C. Park, ``{{PairwiseNet}}: Pairwise collision distance learning for high-dof robot systems,'' in \emph{{{Conference}} on {{Robot Learning}} {(CoRL)}}, 2023.

\bibitem{chasekew2021NeuralCollisionClearance}
J.~Chase~Kew, B.~Ichter, M.~Bandari, T.-W.~E. Lee, and A.~Faust, ``Neural collision clearance estimator for batched motion planning,'' in \emph{Workshop on the Algorithmic {{Foundations}} of {{Robotics}} (WAFR)}, 2021.

\bibitem{ERICSON20057}
C.~Ericson, ``Chapter 2 - {Collision} detection design issues,'' in \emph{Real-Time Collision Detection}.\hskip 1em plus 0.5em minus 0.4em\relax Morgan Kaufmann, 2005, pp. 7--21.

\bibitem{sellan2021SweptVolumesSpacetime}
S.~Sell\'{a}n, N.~Aigerman, and A.~Jacobson, ``Swept volumes via spacetime numerical continuation,'' \emph{ACM Transactions on Graphics}, vol.~40, no.~4, Jul. 2021.

\bibitem{zhang2023ContinuousImplicitSDF}
T.~Zhang, J.~Wang, C.~Xu, A.~Gao, and F.~Gao, ``Continuous implicit {SDF} based any-shape robot trajectory optimization,'' in \emph{{{IEEE}}/{{RSJ International Conference}} on {{Intelligent Robots}} and {{Systems}} ({{IROS}})}, 2023.

\bibitem{wang2024ImplicitSweptVolume}
J.~Wang, T.~Zhang, Q.~Zhang, C.~Zeng, J.~Yu, C.~Xu, L.~Xu, and F.~Gao, ``Implicit swept volume {SDF}: Enabling continuous collision-free trajectory generation for arbitrary shapes,'' \emph{ACM Transactions on Graphics}, vol.~43, no.~4, Jul. 2024.

\bibitem{lin1993efficient}
M.~C. Lin, \emph{Efficient collision detection for animation and robotics}.\hskip 1em plus 0.5em minus 0.4em\relax University of California, Berkeley, 1993.

\bibitem{tang2014HierarchicalControlledAdvancement}
M.~Tang, D.~Manocha, and Y.~J. Kim, ``Hierarchical and controlled advancement for continuous collision detectionof rigid and articulated models,'' \emph{IEEE Transactions on Visualization and Computer Graphics}, vol.~20, no.~5, pp. 755--766, May 2014.

\bibitem{tang2011CCQEfficientLocal}
M.~Tang, Y.~J. Kim, and D.~Manocha, ``{{CCQ}}: Efficient local planning using connection collision query,'' in \emph{Workshop on the Algorithmic {{Foundations}} of {{Robotics}} (WAFR)}, 2011.

\bibitem{Parker2010optix}
S.~G. Parker, J.~Bigler, A.~Dietrich, H.~Friedrich, J.~Hoberock, D.~Luebke, D.~McAllister, M.~McGuire, K.~Morley, A.~Robison, and M.~Stich, ``{OptiX: {A} general purpose ray tracing engine},'' \emph{ACM Transactions on Graphics}, vol.~29, no.~4, Jul. 2010.

\bibitem{dimitrov2009bounds}
D.~Dimitrov, C.~Knauer, K.~Kriegel, and G.~Rote, ``Bounds on the quality of the pca bounding boxes,'' \emph{Computational Geometry}, vol.~42, no.~8, pp. 772--789, 2009.

\bibitem{Pharr2005GPU}
M.~Pharr and R.~Fernando, \emph{GPU Gems 2: Programming Techniques for High-Performance Graphics and General-Purpose Computation}.\hskip 1em plus 0.5em minus 0.4em\relax Addison-Wesley Professional, 2005, ch.~39.

\bibitem{farin2014curves}
G.~Farin, \emph{Curves and Surfaces for Computer-Aided Geometric Design: A Practical Guide}.\hskip 1em plus 0.5em minus 0.4em\relax Elsevier, 2014.

\bibitem{dong2017efficient}
W.~Dong, Y.~Ding, J.~Huang, X.~Zhu, and H.~Ding, ``An efficient approach of time-optimal trajectory generation for the fully autonomous navigation of the quadrotor,'' \emph{Journal of Dynamic Systems, Measurement, and Control}, vol. 139, no.~6, 2017.

\bibitem{crane2013robust}
K.~Crane, U.~Pinkall, and P.~Schr{\"o}der, ``Robust fairing via conformal curvature flow,'' \emph{ACM Transactions on Graphics}, vol.~32, no.~4, pp. 1--10, 2013.

\bibitem{trimesh}
{Dawson-Haggerty et al.}, ``trimesh,'' \url{https://trimesh.org/}.

\bibitem{gpytoolbox}
S.~Sell\'{a}n, O.~Stein \emph{et~al.}, ``{gptyoolbox}: A {Python} {Geometry} {Processing} {Toolbox},'' \url{https://gpytoolbox.org/}.

\bibitem{hoetzlein2016GVDBRaytracingSparse}
R.~K. Hoetzlein, ``{{GVDB}}: Raytracing sparse voxel database structures on the {{GPU}},'' in \emph{{{High Performance Graphics}}}, 2016.

\bibitem{halton1964Algo}
J.~H. Halton, ``Algorithm 247: Radical-inverse quasi-random point sequence,'' \emph{Communications of the ACM}, vol.~7, no.~12, p. 701–702, Dec. 1964.

\end{thebibliography}

\iftoggle{extended_version}{\appendices
\section{Further Implementation Details} \label{app: confs}
We made a number of design decisions as a result of exploration and experimentation to build performant pipelines for collision detection with OptiX and NVIDIA ray-tracing cores. Key algorithm decision points included the following:
\begin{itemize}
    \item Using OptiX geometric acceleration structures (GAS) versus using an instance acceleration structure (IAS) for meshes/curves collision-checked in the same batch
    \item Using versus omitting a custom single-level broad-phase elimination of inputs into the ray-tracing pipeline (either rays or collision geometries)
    \item Choosing sub-batch sizes and the number of asynchronous CUDA streams algorithm sections run on
    \item Optimizing acceleration structure builds for fast builds versus optimizing for fast ray traces against them
\end{itemize}
Since some of these choices are coupled, we will first introduce each concept and then list our recommendations and justifications for relevant collision detection algorithms.
\subsection{Decision Points and Tradeoffs}
\subsubsection{GAS and IAS}
In NVIDIA's OptiX ray-tracing engine, geometric acceleration structures (GAS) and instance acceleration structures (IAS) are two types of acceleration structures which represent geometries to traced rays against (as of this writing, NVIDIA implements these acceleration structures as BVHs using AABB bounding volumes). A GAS consists of only geometric primitives (triangles, spheres, curves, etc.). on the other hand, an IAS is a tree of other acceleration structures which could be GASs or other IASs, each of which has an $SE(3)$ transform relative to its parent and can be included more than once. When geometries are duplicated, an IAS can be more efficient than a GAS, both for build times and memory usage. For example, if a mesh is duplicated across a scene, one can build a single GAS for the mesh and include it multiple times in an IAS in different poses. However, a single large GAS is more free to optimize a single acceleration structure for all the low-level primitives in a scene, which can sometimes lead to faster ray traces.

When tracing against a geometry such as a mesh without duplicated submeshes, there are differences between representing it as a pure GAS vs. a GAS in an IAS when the geometry is given a pose. With a pure GAS, the rays must be manually posed, for example in a ray generation program which runs on CUDA cores. With a posed GAS in an IAS, the pose is applied to the rays during the scene traversal.
Thus we must carefully decide whether to represent the obstacles, batch-posed robots, and trajectories as GASs or IASs to achieve the best performance.

\subsubsection{Custom Single-Level Broad-Phase Elimination}
Although full BVH construction and traversal are complex to implement with high performance on the GPU, a single-level custom broad-phase elimination pass can be very simple while producing significant reductions in inputs to the ray-tracing steps for collision detection. For example, in our experiments, often only $\sim$1/4th of single-level oriented bounding boxes (OBBs) encompassing robot links were in collision with OBBs representing the obstacles.

While the OptiX acceleration structures reduce the number of geometries each ray needs to be traced against, we can use our custom single-level broad-phase to reduce the number of rays traced at all along both the robot and obstacles, which also has a large impact on performance. We can also build simpler acceleration structures containing only robot links which are true collision candidates. However, since the correct links cannot be known ahead of time, this requires building new acceleration structures online rather than performing a fast in-place dynamic update on a pre-built acceleration structure online. 

\subsubsection{Sub-Batch Size and Stream Count}
Given a large batch of collision detection queries, we can process the whole batch together or split it into smaller sub-batches which are each checked using a single ray-tracing call and a single call of other pre- or post-processing CUDA kernels. For example, in DCD ObsToRobot and in our CCD we trace obstacle rays in one large batch, regardless of the number of poses or trajectories checked in the batch.

Increasing the size of one large batch can increase parallelism up to the point when regions of the GPU become saturated and produce bottlenecks. Bottlenecks can occur in different GPU regions, such as in compute (e.g., CUDA cores or RT cores) or in memory, hitting memory access or data bus throughput limits. Since these regions can often perform simultaneous operations with each other, we can sometimes relieve bottlenecks by splitting a large batch into smaller sub-batches. We can then schedule each sub-batch's operations on a separate CUDA stream (a queue of sequential tasks which can execute in parallel with tasks queued on other streams). This provides opportunity for parts of a sub-batch's compute pipeline with heavy use of some GPU region to execute in parallel with a different part of another sub-batch's compute pipeline which doesn't use the region as heavily.

On the other hand, too many small batches on too many streams may incur too much kernel launch and scheduling overhead. Thus, to handle large collision detection batch requests, we could use CUDA streams to divide the batch into sub-batches to maximize device occupancy, but the optimal sub-batch size depends on the algorithm and the particular GPU being used.

\subsubsection{Acceleration Structure Build Flags}
OptiX provides different build options for acceleration structures depending on user priorities. Two of the main options switch between preferring a fast acceleration structure builds versus preferring faster ray traces against the resulting acceleration structures. In most cases, we chose faster ray traces, but in CCD, we must build curve GASs online, making fast build times more important. In particular, for piecewise linear curves, we found prioritizing fast builds resulted in better overall algorithm speeds.

\begin{figure}[t!]
    \centering\includegraphics[width=0.85\columnwidth]{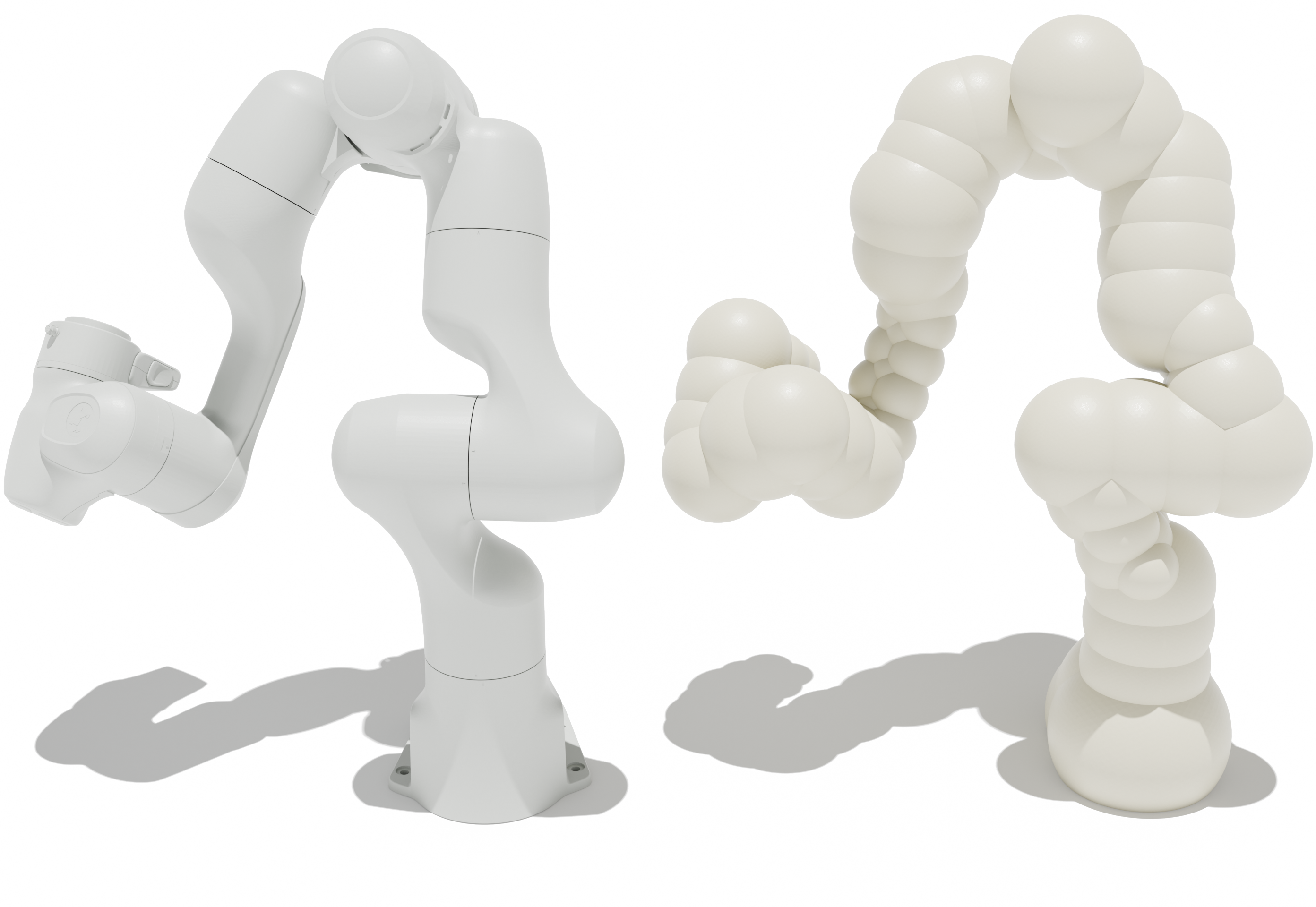}
    \caption{Sphere representation of the Franka Panda robot used in our experiments.}
    \label{fig: sphr_rep}
\end{figure}

\subsection{Choices for Algorithms}
Based on experimentation, we settled on the following configurations. Note that these may be subject to change on different GPUs and on different variations of our algorithms.
\subsubsection{DCD ObsToRobot}
When shooting rays from obstacles to robots, we use an Instance Acceleration Structure (IAS) to represent batched robot links across different robot poses. The IAS saves memory by avoiding duplication of link meshes, which would be needed by a Geometry Acceleration Structure (GAS). We use broad-phase collision detection to filter out any links that can never collide. We then gather selected link instances through stream compaction~\cite{Pharr2005GPU} and use them to build an IAS. On the other side, we found it was rare that obstacles are collision-free across all poses in a batch, so we do not filter them using the broad-phase collision detection results. In our testing, we found it was best to use 4 streams and 1024 poses per batch, which achieves high device occupancy and detection throughput.

\begin{figure*}[t!]
    \centering
    \subfloat[Dense Scene]{\includegraphics[width=0.32\textwidth]{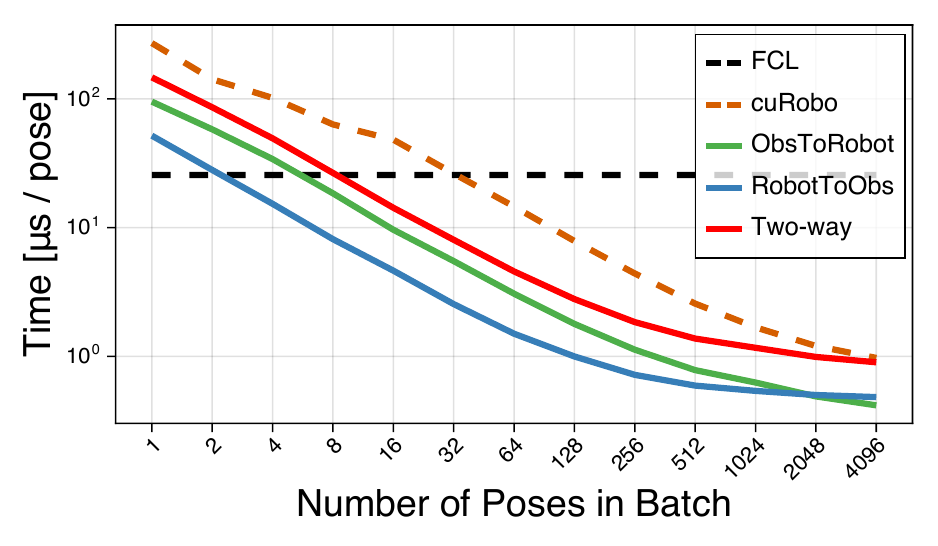}}\
    \subfloat[Medium Scene]{\includegraphics[width=0.32\textwidth]{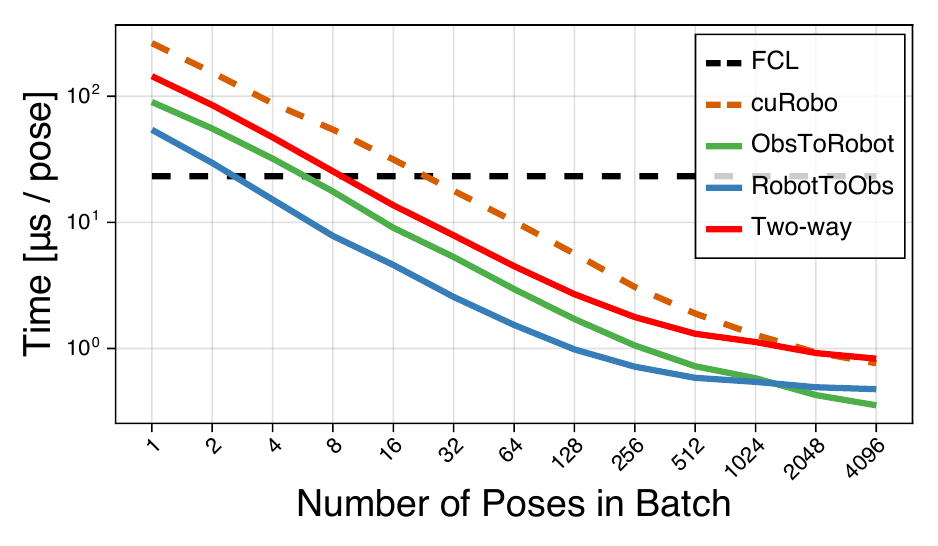}}\
    \subfloat[Simple Scene]{\includegraphics[width=0.32\textwidth]{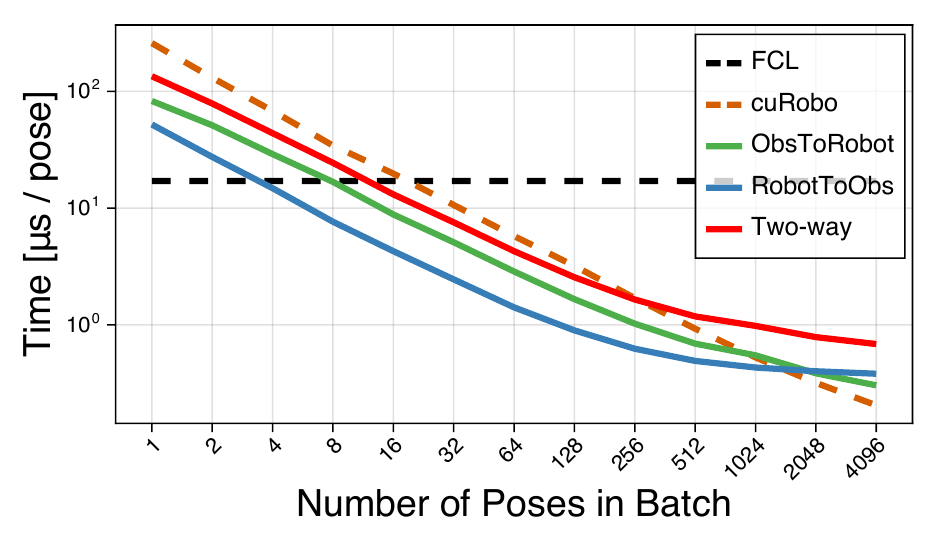}}\
    \caption{Runtimes for discrete collision detection (DCD) algorithms in different scenes with cuRobo CUDA graphs disabled.}
    \label{fig: dcd_cuda_graphs_disabled}
    \vspace{-5pt}
\end{figure*}

\begin{figure*}[t!]
    \centering
    \subfloat[Dense Scene]{\includegraphics[width=0.32\textwidth]{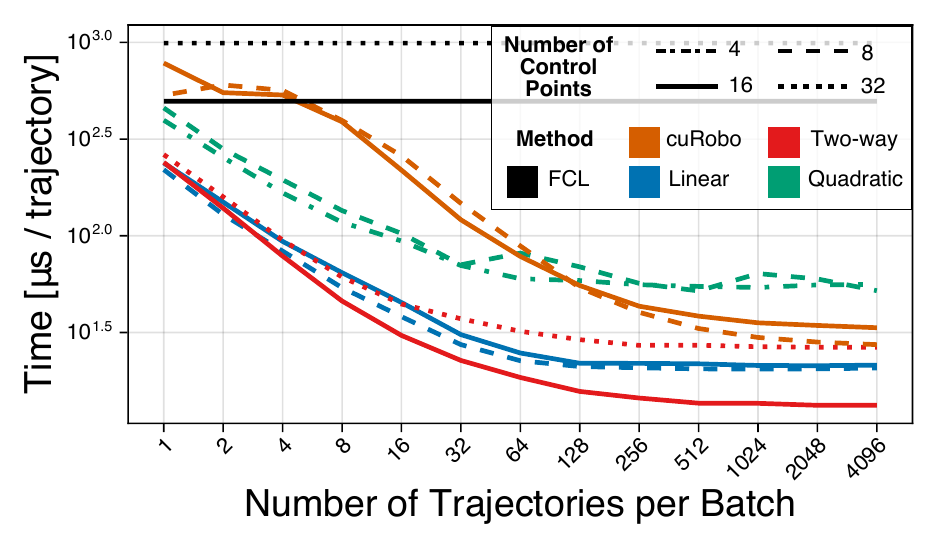}}\
    \subfloat[Medium Scene]{\includegraphics[width=0.32\textwidth]{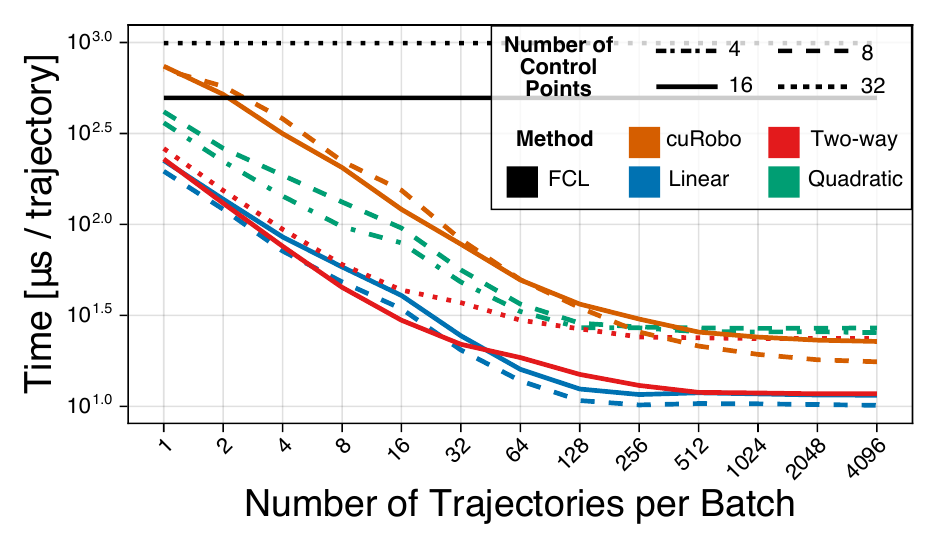}}\
    \subfloat[Simple Scene]{\includegraphics[width=0.32\textwidth]{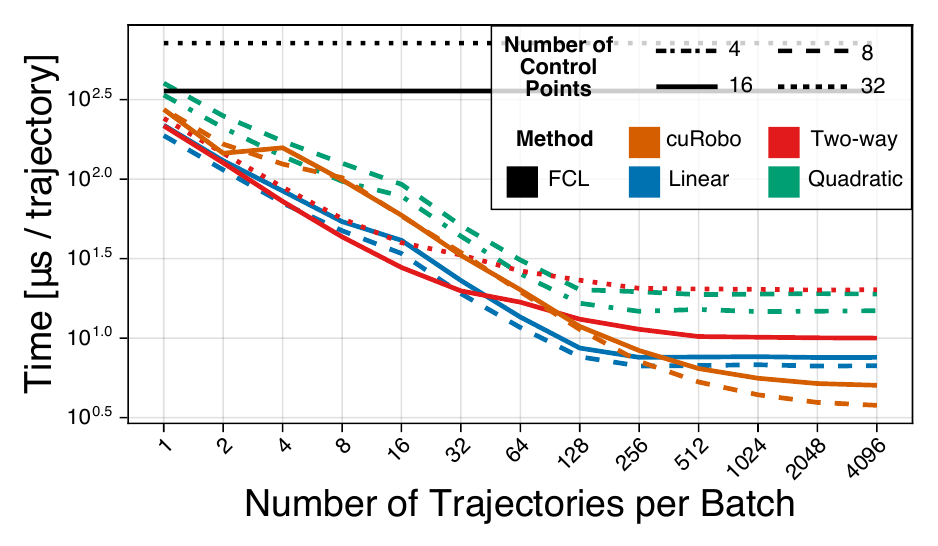}}\
    \caption{Runtimes for continuous collision detection (CCD) algorithms in different scenes with cuRobo CUDA graphs disabled.}
    \label{fig: ccd_cuda_graphs_disabled}
    \vspace{-15pt}
\end{figure*}

\subsubsection{DCD RobotToObs}
For this algorithm, we represent the obstacles using a GAS, which we can build offline and reuse since we have a static obstacle scene. We use the broad-phase collision detection results to collect potential colliding rays along transformed robot links. We observed it is better to transform link rays online rather than offline as it improves cache performance and thereby reduces global memory accesses. As with ObsToRobot, was also found 4 streams and 1024 poses per batch  is the best configuration.

\subsubsection{CCD}
Robot swept volumes change with each trajectory, preventing us from using IASs of pre-built GASs or performing fast dynamic updates of GASs. Instead, we are forced to build a new GAS for each batch of trajectories. We did try a custom curve segment OBB broad-phase to reduce the number of rays traced. However, this did not provide improvement, due to a combination of most of the space being filled with swept volumes and OBBs often being an inefficient bounding volume for swept sphere curves. Thus, the ray-tracing computation saved was outweighed by the extra overhead of running broad-phase collision detection. For this algorithm, we also landed on 4 streams but chose a smaller batch size of 128 trajectories per stream.

\subsection{cuRobo Benchmark Parameters}\label{app: parameters}\vspace{2pt}
In our benchmark experiments, we made slight modifications to the cuRobo library and adjusted inputs to ensure a fair comparison. We adapted our cuRobo benchmark code from the benchmark code provided with the cuRobo library.

First, since our algorithms currently do not include self-collision detection, we disabled cuRobo's self-collision detection. We also replaced cuRobo's sphere representation of the Franka Panda robot with the same one used for our own algorithm. This sphere set has a few more spheres with less conservatism (shown in Fig.~\ref{fig: collision scene} and~\ref{fig: sphr_rep}). For the CCD experiments, we set the cuRobo trajectory horizon to 8 and 16 to match our number of curve spline control points. We also set cuRobo's \texttt{sweep\_step} to 3, which gives it enough opportunity to capture collisions in dense obstacle environments.

We enabled CUDA graphs in all benchmarks to maximize cuRobo's performance, resulting in a slight speedup and different trends across batch sizes. For comparison, the results with CUDA graphs disabled in cuRobo are shown in Fig.~\ref{fig: dcd_cuda_graphs_disabled} and~\ref{fig: ccd_cuda_graphs_disabled}.

\subsection{CCD Strongly-Connected Directed Graph Construction}\label{app: parameters}
As explained in Sec.~\ref{subsec: CCD}, to avoid tracing two rays along each obstacle mesh edge in CCD, we build a strongly-connected directed graph on the each obstacle mesh's edges and trace only along these directed graph edges, significantly reducing the number of rays required without sacrificing correctness. Due to limited space in the main paper, the explanation of our simple approach to construct this graph was quite brief, so we provide the pseudocode in Alg.~\ref{alg: buildLoopEdges}.

\begin{algorithm}[]
    \caption{Constructing a Strongly-Connected Directed Graph on Connected Triangle Mesh Edges}
    \label{alg: buildLoopEdges}
    \textbf{Input:} Mesh triangles $T$, edge-to-triangle map \\
    \textbf{Output:} Mesh edge directions
    \begin{algorithmic}[1]
        \State $T_\text{processed} \gets \emptyset \quad \blacktriangleright$ Processed triangles
        \State $t \gets$ randomly select initial triangle from $T$
        \State $T_\text{wavefront} \gets \{t\} \quad \blacktriangleright$ Triangles on expansion wavefront
        \State $T_\text{unprocessed} \gets T \setminus \{t\} \quad \blacktriangleright$ Unprocessed triangles
        
        \While{$T_\text{wavefront} \neq \emptyset$}
            \State $t \gets$ Sample and remove triangle from $T_\text{wavefront}$
            \State $T_\text{neighbors} \gets$ neighbors of $t$ in $T_\text{processed}$
            \State $n \gets |T_\text{neighbors}| \;\; \blacktriangleright$ \# of neighbor triangles (up to 3)
            \If{$n = 0$ or $T_\text{neighbors}$ orientations are tied}
                \State Set $t$ to random orientation$\quad\blacktriangleright n = 0$ or $2$
            \ElsIf{all $T_\text{neighbors}$ orientations are the same}
                \State Set $t$ to opposite orientation $\quad\blacktriangleright n = 1, 2$, or $3$ 
            \Else
                \State Set $t$ orientation opposite to majority$\quad\blacktriangleright n = 3$ 
            \EndIf
            \State $T_\text{processed} \gets T_\text{processed}\bigcup\{t\}$
            \State Move neighbors of $t$ in $T_\text{unprocessed}$ to $T_\text{wavefront}$
        \EndWhile

        \For{edge in mesh}
            \For{triangle along edge}
                \State Add edge direction matching triangle orientation
            \EndFor
        \EndFor
        \State \Return edge directions
    \end{algorithmic}
\end{algorithm}
}{}

\end{document}